\def\year{2022}\relax
\documentclass[letterpaper]{article} %
\usepackage{xcolor}
\usepackage{aaai22}  %
\usepackage{times}  %
\usepackage{helvet}  %
\usepackage{courier}  %
\usepackage[hyphens]{url}  %
\usepackage{graphicx} %
\usepackage{natbib}  %
\usepackage{caption} %
\DeclareCaptionStyle{ruled}{labelfont=normalfont,labelsep=colon,strut=off} %
\usepackage{hyperref}
\usepackage{enumitem}
\usepackage[ruled, vlined]{algorithm2e} %
\usepackage{algpseudocode}
\usepackage{forloop}

\setlist[enumerate]{label=(\alph*),itemsep=2pt,topsep=0pt,leftmargin=8mm} %
\renewcommand{\paragraph}[1]{\vspace{2pt}\noindent{\textbf{#1}}}

\title{Model-based Offline Reinforcement Learning with Local Misspecification}
\author{
	Kefan Dong\equalcontrib,
	Yannis Flet-Berliac\equalcontrib, 
	Allen Nie\equalcontrib,
    Emma Brunskill
   	
}
\affiliations{

	Stanford University\\ \{kefandong,yfletberliac,anie,ebrun\}@stanford.edu
}

\usepackage{booktabs}
\usepackage{amsmath,amsfonts,amsthm,amssymb}
\usepackage{pifont}
\usepackage{caption}
\usepackage{subcaption}
\newcommand{\cmark}{\ding{51}}
\newcommand{\xmark}{\ding{55}}
\allowdisplaybreaks[1]

\newtheorem{theorem}{Theorem}
\newtheorem{lemma}[theorem]{Lemma}
\newtheorem{proposition}[theorem]{Proposition}

\newtheorem{definition}[theorem]{Definition}

\usepackage{amsmath,amsfonts,bm}

\newcommand*{\defeq}{\triangleq}

\def\1{\bm{1}}

\makeatletter
\newcommand{\ve}{\@ifnextchar\bgroup{\velong}{{\bm{e}}}}
\newcommand{\velong}[1]{{\bm{#1}}}
\makeatother

\DeclareMathAlphabet{\mathsfit}{\encodingdefault}{\sfdefault}{m}{sl}
\SetMathAlphabet{\mathsfit}{bold}{\encodingdefault}{\sfdefault}{bx}{n}

\def\calA{{\mathcal{A}}}

\def\calD{{\mathcal{D}}}
\def\calE{{\mathcal{E}}}

\def\calG{{\mathcal{G}}}
\def\calH{{\mathcal{H}}}

\def\calN{{\mathcal{N}}}
\def\calO{{\mathcal{O}}}

\def\calS{{\mathcal{S}}}
\def\calT{{\mathcal{T}}}

\def\calW{{\mathcal{W}}}
\def\calX{{\mathcal{X}}}

\newcommand{\E}{\mathbb{E}}
\newcommand{\R}{\mathbb{R}}

\DeclareMathOperator*{\argmax}{argmax}
\DeclareMathOperator*{\argmin}{argmin}

\newcommand{\poly}{\mathrm{poly}}

\newcommand{\dotp}[2]{\left<#1, #2\right>}

\newcommand{\norm}[1]{\left\| #1 \right\|}

\newcommand{\normone}[1]{\left\| #1 \right\|_1}

\newcommand{\pbra}[1]{{\left( {#1} \right)}}
\newcommand{\bbra}[1]{{\left[ {#1} \right]}}

\newcommand{\bbrasm}[1]{{\Bigl[ {#1} \Bigr]}}
\newcommand{\abs}[1]{{\left| {#1} \right|}}
\newcommand{\abssm}[1]{{\Bigl| {#1} \Bigr|}}

\newcommand{\ind}[1]{\mathbb{I}\left[ #1 \right]}
\newcommand{\indsm}[1]{\mathbb{I}\Bigl[ #1 \Bigr]}

\newcommand{\bigO}{\calO}
\newcommand{\tildeO}{\widetilde{\calO}}

\newcommand{\dy}{T}
\newcommand{\lb}{\mathrm{lb}}
\newcommand{\caldy}{\calT}
\newcommand{\dystar}{{T^\star}}
\newcommand{\valstar}{V^\pi_{\dy^\star}}

\newcommand{\hdy}{{\hat{\dy}}}
\newcommand{\hpi}{{\hat{\pi}}}
\newcommand{\hE}{{\hat{\E}}}

\newcommand{\TV}[2]{\mathrm{TV}\left( #1, #2 \right)}
\newcommand{\vmax}{V_{\max}}
\newcommand{\rmax}{R_{\max}}

\newcommand{\hmu}{\hat\mu}

\begin{document}

\maketitle

\begin{abstract}
We present a model-based offline reinforcement learning policy performance lower bound that explicitly captures dynamics model misspecification and distribution mismatch and we propose an empirical algorithm for optimal offline policy selection. Theoretically, we prove a novel safe policy improvement theorem by establishing pessimism approximations to the value function. Our key insight is to jointly consider selecting over dynamics models and policies: 
as long as a dynamics model can accurately represent the dynamics of the state-action pairs \emph{visited} by a given policy, it is possible to approximate the value of that particular policy. We analyze our lower bound in the LQR setting and also show competitive performance to previous lower bounds on policy selection across a set of D4RL tasks. 
\end{abstract}

\section{Introduction}\label{sec:intro}

Offline reinforcement learning (RL) could 
leverage historical decisions made and their outcomes to improve data-driven decision-making in areas like marketing \citep{thomas2017predictive}, robotics \citep{quillen2018deep,yu2020mopo,yu2021combo,swazinna2020overcoming, singh2020cog}, recommendation systems \citep{swaminathan2015batch}, etc. Offline RL is particularly useful when it is possible to 
deploy context-specific decision policies, but it is costly or infeasible to do online reinforcement learning.

Prior work on offline RL for large state and/or action spaces has primarily focused on one of two extreme settings. One line of work makes minimal assumptions on the underlying stochastic process, requiring only no confounding, and leverages importance-sampling estimators of potential policies (e.g.,~\citet{thomas2015high,thomas2019preventing}). Unfortunately, such estimators have a variance that scales exponentially with the horizon \citep{liu2018representation} and are often ill-suited to long horizon problems\footnote{Marginalized importance sampling (MIS) methods \citep{liu2018breaking,xie2019towards,yin2020asymptotically,liu2020understanding} help address this but rely on the system being Markov in the underlying state space}. 

An alternative, which is the majority of work in offline RL, is to make a number of assumptions on the domain, behavior data generation process and the expressiveness of the function classes employed. The work in this space typically assumes the domain satisfies the Markov assumption, which has been recently shown in the off-policy evaluation setting to enable provably more efficient policy value estimation \cite{kallus2020double}. Historically, most work (e.g.,~\citet{munos2003error,farahmand2010error, xie2020batch,chen2019information}) assumes the batch data set has coverage on any state-action pairs that could be visited under any possible policy. More recent work relaxes this strong requirement using a pessimism under uncertainty approach that is model-based \citep{yu2020mopo,yu2021combo,kidambi2020morel}, model-free \citep{liu2020provably} or uses policy search  \citep{curi2020efficient,van2019use}. 
Such work still relies on realizability/lack of misspecification assumptions. For model-free approaches, a common assumption is that the value function class can represent all policies.  \citet{liu2020provably} assume that the value function class is closed  under (modified) Bellman backups. A recent exception is  \citet{xie2020batch}, which only requires the optimal $Q$-function to be representable by the value function class. However, their sample complexity scales non-optimally \citep[Theorem 2]{xie2020batch}, and they also make strong assumptions on the data coverage -- essentially the dataset must visit all states with sufficient probability. Model-based approaches such as \citet{malik2019calibrated,yu2020mopo} assume the dynamics class has no misspecification. 

These two lines of work hint at possibilities in the middle: can we leverage the sample-efficient benefits of Markov structure and  allow for minimal assumptions on the data-gathering process and potential model misspecification? This can be viewed as one step towards more best-in-class results for offline RL. Such results are relatively rare in RL, which tends to focus on obtaining optimal or near-optimal policies for the underlying domain. Yet in many important applications, it may be much more practical to hope to identify a strong policy within a particular policy class. 

Our insight is that the algorithm may be able to leverage misspecified models and still leverage the Markov assumption for increased data efficiency. In particular, we take a model-based offline RL approach to leverage dynamics models that can accurately fit the space of state-action pairs visited under a particular policy (\textit{local small misspecification}), rather than being a good model of the entire possible state-action space (\textit{global small misspecification}). 
Our work is most closely related to the recently proposed Minimax Model Learning (MML) algorithm~\citep{voloshin2021minimax}: MML optimizes for the model that minimizes a value-aware error which upper bounds the difference of policy value in learned and real models. If the considered model class includes the true model, this can work very well, but when the models are misspecified, this can become overly conservative since it optimizes with respect to a worst-case potential state-action distribution shift.

The key feature of our algorithm is to jointly optimize policy and dynamics. 
Prior model-based offline RL algorithms typically estimate dynamics first, and then optimize a policy w.r.t. the learned dynamics \citep{yu2020mopo,yu2021combo,voloshin2021minimax}.
But when the dynamics model class is misspecified, there may not exist a unique ``good dynamics'' that can approximate the value of every policy. As a result, the learned policy may have a good estimated value under the learned dynamics, but a poor performance in the real environment, or the learned policy may be overly conservative due to the misestimated dynamics. 

Our paper makes the following contributions. First, we provide a finite sample bound that assumes a Markov model, leverages the pessimism principle to work with many data-gathering distributions, accounts for estimation error in the behavior policy and, most importantly, directly accounts for dynamics and value function model misspecification (see Lemma~\ref{lem:OPE-lowerbound}). We prove the misspecification error of our method is much tighter than other approaches because we only look at the models' ability to represent visited state-action pairs for a particular policy. 
In that sense, we say our algorithm relies on small \emph{local} model dynamics misspecification. In Theorem~\ref{thm:hardinstance}, we show that when the dynamics model class does not satisfy realizability, decoupling the learning of policy and dynamics is suboptimal. This motivates our algorithm which jointly optimizes the policy and model dynamics across a finite set.  
Because of the tighter pessimistic estimation, we can prove a novel safe policy improvement theorem (see Theorem~\ref{thm:OPE-main}) for offline policy optimization (OPO). While our primary contribution is theoretical, our proposed method for policy selection improves over the state-of-the-art MML~\citet{voloshin2021minimax} in a simple linear Gaussian setting, and has solid performance on policy selection on a set of D4RL benchmarks.

\section{Related Works}\label{sec:relatedworks}

There is an extensive and growing body of research on offline RL and we focus here on methods that also assume a Markov domain. Many papers focus on model-free methods (e.g., \citet{fujimoto2018addressing,kumar2019stabilizing,kumarNeurips2020}). 
\citet{nachum2019dualdice} and their follow-ups \citep{zhang2019gendice,zhang2020gradientdice} learn a distribution correction term, on top of which they perform evaluation or policy optimization tasks. \citet{uehara2020minimax,jiang2020minimax} study the duality between learning $Q$-functions and learning importance weights. \citet{liu2020provably} explicitly consider the distribution shift in offline RL and propose conservative Bellman equations. 

Another line of research uses model-based methods \citep{kidambi2020morel,yu2020mopo,yu2021combo,matsushima2020deployment,swazinna2020overcoming,fu2021offline, farahmand2017value}. \citet{gelada2019deepmdp,delgrange2022distillation,voloshin2021minimax} learn the dynamics using different loss functions. \citet{yu2020mopo} build an uncertainty quantification on top of the learned dynamics and select a policy that optimizes the lower confidence bound. \citep{argenson2020model,zhan2021model} focus on policy optimization instead of model learning.

In Table~\ref{table:statistical}, we compare our error bounds with existing results. Our statistical error (introduced by finite dataset) is comparable with VAML \citep{farahmand2017value}, MBS-PI \citep{liu2020provably} and MML \citep{voloshin2021minimax}. In addition, we consider misspecification errors and safe policy improvement (SPI).

\begin{table}[h]
	\begin{center}
		\begin{tabular}{ |c | c | c | c| } 
			\hline
			Algorithm & Statistical Error & Misspecification & SPI\\
			\hline
			VAML &$\tildeO\pbra{\frac{p}{\sqrt{n}}}$\footnotemark[2] & \cmark (global) & \xmark \\
			\hline
			MBS-PI &$\tildeO\pbra{\frac{\vmax\zeta}{(1-\gamma)^{2}\sqrt{n}}}$ & \cmark (global) & \cmark \\
			\hline
			MML &$\mathfrak{R}_n$\footnotemark[3] & \cmark (global) & \xmark \\
			\hline
			Ours & $\tildeO\pbra{\frac{\vmax}{1-\gamma}\sqrt{\frac{\zeta}{n}}}$ & \cmark (local) & \cmark \\
			\hline
		\end{tabular}
	\end{center}
\caption{Comparison of error bounds with prior works.}
\label{table:statistical}
\end{table}

\footnotetext[2]{VAML only considers linear function approximation and $p$ is the dimension of the feature vector.}
\footnotetext[3]{The Rademacher complexity. For the finite hypothesis, the best-known upper bound is in the same order of ours.}

\section{Problem Setup}\label{sec:setup}
A Markov Decision Process (MDP) is defined by a tuple $\left<\dy,r,\calS,\calA,\gamma\right>.$ $\calS$ and $\calA$ denote the state and action spaces. $\dy:\calS\times\calA\to\Delta(\calS)$ is the transition and $r:\calS\times\calA\to\R_+$ is the reward.
$\gamma\in[0,1)$ is the discount factor. For a policy $\pi:\calS\to\Delta(\calA)$, the value function is defined as 
\begin{align*}
	V^\pi_\dy(s)=\E_{s_0=s,a_t\sim\pi(s_t),s_{t+1}\sim \dy(s_t,a_t)}\bbra{\textstyle\sum\nolimits_{t=0}^{\infty}\gamma^tr(s_t,a_t)}.
\end{align*}
Let $\rmax\defeq \max_{s,a}r(s,a)$ be the maximal reward and $\vmax\defeq\rmax/(1-\gamma).$
Without loss of generality, we assume that the initial state is fixed as $s_0$. We use $\eta(\dy,\pi)\defeq V^\pi_\dy(s_0)$ to denote the expected value of policy $\pi$. Let $\rho^\pi_\dy(s,a)\defeq (1-\gamma)\sum_{t=0}^{\infty} \gamma^t \Pr^\pi_T(s_t=s, a_t=a \mid  s_0)$ be the normalized state-action distribution when we execute policy $\pi$ in a domain with dynamics model $\dy.$ For simplicity in this paper we assume the reward function is known. 

An offline RL algorithm takes a dataset $\calD=\{(s_i,a_i,s_i')\}_{i=1}^{n}$ as input, where $n$ is the size of the dataset. Each $(s_i,a_i,s_i')$ tuple is drawn independently from a behavior distribution $\mu.$ We assume that $\mu$ is consistent with the MDP in the sense that $\mu(\cdot\mid s,a)=\dy(s,a)$ for all $(s,a).$ For simplicity, we use $\hat{\E}$ to denote the empirical distribution over the dataset $\calD$. In this paper, we assume that the algorithm has access to an estimated behavior distribution $\hmu$ such that $\mathrm{TV}(\mu,\hmu)$ is small. This estimation can be easily obtained using a separate dataset (e.g., \citet{liu2020provably}).

The algorithm can access three (finite) function classes $\calG,\calT,\Pi$. $\calG$ is a class of value functions, $\calT$ a class of dynamics and $\Pi$ a class of policies. We assume that $g(s,a)\in[0,\vmax]$ for all $g\in\calG$. We use $\dystar$ to denote the ground-truth dynamics. Note that $\dystar$ may not be in $\calT$. Our goal is to return a policy $\pi\in\Pi$ that maximizes $\eta(\dystar,\pi)$.
\section{Main Results}\label{sec:main} %
A standard model-based RL algorithm learns the dynamics models first, and then uses the learned dynamics to estimate the value of a policy, or optimize it. In this approach, it is crucial to link the estimation error of the dynamics to the estimation error of the value. Therefore, as a starting point, we invoke the simulation lemma.
\begin{lemma}[Simulation Lemma \citep{yu2020mopo,kakade2002approximately}]\label{lem:simulation}
	Consider two MDPs with dynamics $\dy,\dystar$, and the same reward function. Then,
	\begin{align}
		&\eta(\dy,\pi)-\eta(\dystar,\pi)=\frac{\gamma}{1-\gamma}\E_{(s,a)\sim \rho^\pi_\dy}\left[\right.\nonumber\\
		&\quad \left.\E_{s'\sim \dy(s,a)}[V^\pi_{\dystar}(s')]-\E_{s'\sim \dystar(s,a)}[V^\pi_{\dystar}(s')]\right].
	\end{align}
\end{lemma}
For a fixed ground-truth dynamics $\dystar$, we define $G^\pi_\dy(s,a)=\E_{s'\sim \dy(s,a)}[V^\pi_{\dystar}(s')]-\E_{s'\sim \dystar(s,a)}[V^\pi_{\dystar}(s')]$. The simulation lemma states that the dynamics will provide an accurate estimate of the policy value  if $\E_{s'\sim \dy(s,a)}[V^\pi_{\dystar}(s')]$ matches $\E_{s'\sim \dystar(s,a)}[V^\pi_{\dystar}(s')]$. In other words, to obtain a good estimate of a policy value, it is sufficient to minimize the model error $G^\pi_\dy(s,a)$. 

Since the value function $V^\pi_{\dystar}$ is unknown, \citet{yu2020mopo} upper bound the model error by introducing a class of test functions $\calG: \calS\to\R$.
When $V^\pi_{\dystar}\in \calG$, we have
\begin{align*}%
	\textstyle{\abs{G^\pi_\dy(s,\!a)}\!  \le \! \sup_{g\in\calG}\abs{\E_{s'\sim\dy(s,a)}g(s')\!-\!\E_{s'\sim\dystar(s,a)}g(s')]}.}
\end{align*}
In an offline dataset $\calD$, typically we can only observe one sample from $\dystar(s,a)$ per state-action pair. Hence the algorithm cannot compute this upper bound exactly. In addition, the distribution of the dataset $\calD$ is also different from the one required by the simulation lemma $\rho^\pi_\dy$. To address these issues, we explicitly introduce a density ratio $w:\calS\times\calA\to \R_+$. For a test function $g\in \calG$ and a dynamics model $\dy$, let $f^g_\dy(s,a)\defeq\E_{s'\sim \dy(s,a)}[g(s')]$. Recall that $\hE$ denotes the empirical expectation over dataset $\calD$. Then our model loss is defined as
\begin{align}\label{equ:loss}
	\textstyle{\ell_w(\dy,g)=|\hat{\E}\bbra{w(s,a)\pbra{f^g_{\dy}(s,a)-g(s')}}|.}
\end{align}

\paragraph{Distribution mismatch.}
We aim to upper bound policy evaluation error by the loss function even if there are state action pairs with small probability mass under behavior distribution $\mu$ (i.e., the offline dataset does not have a perfect coverage). Following \citet{liu2020provably}, we treat the unknown state-action pairs pessimistically. Let $\zeta$ be a fixed cutoff threshold. Recall that $\hmu$ is an estimation of the behavior distribution. For a policy $\pi$ and dynamics $\dy$, we define $w_{\pi,\dy}(s,a)\defeq\ind{\frac{\rho^\pi_\dy(s,a)}{\hmu(s,a)}\le \zeta}\frac{\rho^\pi_\dy(s,a)}{\hmu(s,a)}$ as the truncated density ratio. For a fixed policy $\pi$, when $w=w_{\pi,\dy},$
\begin{align*}
	&\abssm{\E_{(s,a)\sim \rho^\pi_\dy}\bbrasm{G^\pi_\dy(s,a)}}\\
	\le\;&\abs{\E_{(s,a)\sim \rho^\pi_\dy}\bbrasm{\indsm{\frac{\rho^\pi_\dy(s,a)}{\hmu(s,a)}\le \zeta}G^\pi_\dy(s,a)}}\\
	&\quad +\abssm{\E_{(s,a)\sim \rho^\pi_\dy}\bbrasm{\indsm{\frac{\rho^\pi_\dy(s,a)}{\hmu(s,a)}> \zeta}G^\pi_\dy(s,a)}}\\
	\le\;&|\E_{(s,a)\sim \hmu}\bbrasm{w(s,a)G^\pi_\dy(s,a)}|\\
	&\quad +\vmax\abssm{\E_{(s,a)\sim \rho^\pi_\dy}\bbrasm{\indsm{\frac{\rho^\pi_\dy(s,a)}{\hmu(s,a)}> \zeta}}}\\
	\le\;&|\E_{(s,a)\sim \mu}\bbrasm{w(s,a)G^\pi_\dy(s,a)}|+\zeta\vmax\TV{\hmu}{\mu}\\
	&\quad +\vmax\abs{\E_{(s,a)\sim \rho^\pi_\dy}\bbrasm{\indsm{\frac{\rho^\pi_\dy(s,a)}{\hmu(s,a)}> \zeta}}}.
\end{align*}
Hence, ignoring statistical error due to finite dataset, we can upper bound the estimation error $|\eta(\dy^\star,\pi)-\eta(\dy,\pi)|$ by
\begin{align}\label{equ:lowerbound}
	\frac{\gamma}{1-\gamma}\Big(&\sup_{g\in \calG}\abssm{\ell_{w_{\pi,\dy}}(g,\dy)}+\zeta\vmax\TV{\hmu}{\mu}\nonumber\\
	&\quad+\vmax\E_{(s,a)\sim\rho^\pi_\dy}\bbrasm{\indsm{\frac{\rho^\pi_\dy(s,a)}{\hat\mu(s,a)}> \zeta}}\Big).
\end{align}
Intuitively, the first term measures the error caused by imperfect dynamics $\dy$, the second term captures the estimation error of the behavior distribution, and the last term comes from truncating the density ratios. 

\subsection{Pessimistic Policy Optimization with Model Misspecification}
In this section, we explicitly consider misspecifications of the function classes used for representing the value function and dynamics models ($\calG$ and $\caldy,$ respectively). Most prior theoretical work on model-based RL make strong assumptions on the realizability of the dynamics model class. For example, in the offline setting, \citet{voloshin2021minimax} focus on exact realizability of the dynamics model (that is, $\dystar\in \caldy$). In the online setting, \citet{jin2019provably} provide bounds where there is a linear regret term due to global model misspecification. Their bounds require a $\dy\in \caldy$ such that $\TV{\dy(s,a)}{\dystar(s,a)}\le \epsilon$ for all $(s,a)$, even if the state-action pair $(s,a)$ is only visited under some poorly performing policies. We now show that offline RL tasks can need much weaker realizability assumptions on the dynamics model class.

Our key observation is that for a given dynamics $\dy$ and policy $\pi$, computing the density ratio $w_{\pi,\dy}$ is \emph{statistically} efficient. Note that to compute $w_{\pi,\dy}$ we do not need any samples from the true dynamics: instead, we only need to be able to estimate the state-action density under a dynamics model $\dy$ for policy $\pi$. This allows us to explicitly utilize the density ratio to get a relaxed realizability assumption. 
\begin{definition}\label{def:local-mis}
	The local value function error for a particular dynamics model $\dy$ and policy $\pi$ is defined as 
	\begin{align*}
		&\epsilon_V(\dy,\pi)\defeq \inf_{g\in \calG}|\E_{(s,a)\sim \mu}[w_{\pi,\dy}(s,a)(\E_{s'\sim \dy(s,a)}[(g-V^\pi_\dystar)(s')]\\
		&\quad+\E_{s'\sim \dystar(s,a)}[(g-V^\pi_\dystar)(s')])]|.
	\end{align*}
\end{definition}
The term $\epsilon_V$ measures the local misspecification of the value function class -- that is, the error between the true value of the policy $V^\pi_\dystar$ and the best fitting value function in the class $\calG$ -- \emph{only} on  
the state-action pairs that policy $\pi$  visits under a particular potential dynamics model $\dy$. In contrast, previous results \citep{jin2019provably,nachum2019dualdice,voloshin2021minimax} take the global maximum error over all (reachable) $(s,a)$, which can be much larger than the local misspecification error $\epsilon_V(\dy,\pi)$.

With this local misspecification error, we can establish a pessimistic estimation of the true reward. Let $\calE$ be a high probability event under which the loss function $\ell_{w_{\pi,\dy}}(\dy,g)$ is close to its expectation (randomness comes from the dataset $\calD$). In the Appendix, we define this event formally and prove that $\Pr(\calE)\ge 1-\delta.$ The following lemma gives a lower bound on the true reward. Proofs, when omitted, are in the Appendix.
\begin{lemma}\label{lem:OPE-lowerbound}
	Let $\iota= \log(2|\calG||\caldy||\Pi|/\delta)$. For any dynamics model $\dy$ and policy $\pi$, we define 
	\begin{align}\label{equ:alg-local-lower-bound}
		&\lb(\dy,\pi)=\eta(\dy,\pi)-\frac{1}{1-\gamma}\Bigl(\sup_{g\in \calG}\ell_{w_{\pi,\dy}}(g,\dy)\nonumber \\
			&\qquad +\vmax\E_{(s,a)\sim\rho^\pi_\dy}\bbrasm{\indsm{\frac{\rho^\pi_\dy(s,a)}{\hat\mu(s,a)}> \zeta}}\Bigr).
	\end{align}
	Then, under the event $\calE$, we have
	\begin{align}\label{equ:lowerbound-mis}
		\eta(\dy^\star,\pi)
		&\ge
		\lb(\dy,\pi)-\frac{1}{1-\gamma}\Bigl(\epsilon_V(\dy,\pi)\nonumber\\
		&-2\vmax\sqrt{\zeta\iota/n}-\zeta\vmax\TV{\hmu}{\mu}\Bigr).
	\end{align}
\end{lemma}
We use this to define our offline policy selection Alg.~\ref{alg:OPE}.

\begin{algorithm}[h]
\SetAlgoLined
	\caption{Model-based Offline RL with Local Misspecification Error}
	\label{alg:OPE}
		 \textbf{Require:} estimated behavior distribution $\hmu$, truncation threshold $\zeta$. \\
		\For{$\pi\in \Pi,\dy\in \caldy$}{
    		 Compute $w_{\pi,\dy}(s,a)=\ind{\frac{\rho^\pi_\dy(s,a)}{\hmu(s,a)}\le \zeta}\frac{\rho^\pi_\dy(s,a)}{\hmu(s,a)}.$ \\
    		 Compute  $\lb(\dy,\pi)$ by Eq.~\eqref{equ:alg-local-lower-bound}. \\
		}
		$\pi\gets \argmax_{\pi\in \Pi}\max_{\dy\in\caldy}\lb(\dy,\pi).$ \\
\end{algorithm}
In contrast to existing offline model-based algorithms \citep{yu2020mopo,voloshin2021minimax}, our algorithm optimizes the dynamics and policy jointly. For a given dynamics model, policy pair, our Alg.~\ref{alg:OPE} computes the  truncated density ratio $w_{\pi,\dy}$ which does not require collecting new samples and then uses this to compute a lower bound $\lb(\dy,\pi)$ (Eq.~\eqref{equ:alg-local-lower-bound}). Finally, it outputs a policy that maximizes the lower bound. We will shortly show this joint optimization can lead to better offline learning.

Parameter $\zeta$ controls the truncation of the stationary importance weights. 
Increasing $\zeta$ decreases the last term in the lower bound objective $\lb(\dy,\pi)$, but it may also increase the variance given the finite dataset size.
Note that by setting 
$\zeta=\log(n)$ and letting $n\to\infty$ (i.e., with infinite data), the last term in Eq.~\eqref{equ:alg-local-lower-bound} and the statistical error converge to zero.

\subsection{Safe Policy Improvement}
We now derive a novel safe policy improvement result, up to the error terms given below. Intuitively this guarantees that the policy returned by Alg.~\ref{alg:OPE} will improve over the behavior policy when possible, which is an attractive property in many applied settings. Note that recent work~\citep{voloshin2021minimax,yu2020mopo} on model-based offline RL does not provide this guarantee when the dynamics model class is misspecified. 
For a fixed policy $\pi$, define
\begin{align}
	&\textstyle{\epsilon_\rho(\pi)\defeq \inf_{\dy\in\caldy}\E_{(s,a)\sim \rho^\pi_\dystar}[\TV{\dy(s,a)}{\dystar(s,a)}]},\label{equ:error-1}\\
	&\epsilon_{\mu}(\pi)\defeq \E_{(s,a)\sim \rho^{\pi}_{\dystar}}\bbrasm{\indsm{\frac{\rho^{\pi}_{\dystar}(s,a)}{\hat\mu(s,a)}>\zeta/2}}.\label{equ:error-2}
\end{align}
The term $\epsilon_\rho$ measures the local misspecification error of the dynamics model class in being able to represent the dynamics for state-action pairs encountered for policy $\pi$. $\epsilon_\mu$ represents that overlap of the dataset for an alternate policy $\pi$: such a quantity is common in much of offline RL. %
In the following theorem, we prove that the true value of the policy computed by Alg.~\ref{alg:OPE} is lower bounded by that of the optimal policy in the function class with some error terms.

\begin{theorem}\label{thm:OPE-main}
	Consider a fixed parameter $\zeta.$ Let $\hpi$ be the policy computed by Alg.~\ref{alg:OPE} and $\hdy=\argmax_{\dy}\lb(\dy,\hpi)$. Let $\iota= \log(2|\calG||\caldy||\Pi|/\delta)$. Then, with probability at least $1-\delta$, we have
	\begin{align}
		&\eta(\dy^\star,\hpi)\ge \sup_{\pi}\left\{\eta(\dy^\star,\pi)-\frac{6\vmax\epsilon_\rho(\pi)}{(1-\gamma)^2}-\frac{\vmax\epsilon_\mu(\pi)}{1-\gamma}\right\}\nonumber\\
		&\quad -\frac{\epsilon_V(\hdy,\hpi)}{1-\gamma}-\frac{4\vmax}{1-\gamma}\sqrt{\frac{\zeta\iota}{n}}-\frac{2\zeta\vmax\TV{\hmu}{\mu}}{1-\gamma}.
	\end{align}
\end{theorem} 

To prove Theorem \ref{thm:OPE-main}, we prove the tightness of $\lb(\dy,\pi)$ --- the lower bound $\max_\dy \lb(\dy,\pi)$ is at least as high as the true value of the policy with some errors. Consequently, maximizing the lower bound also maximizes the true value of the policy. Formally speaking, we have the following Lemma.
\begin{lemma}\label{lem:OPE-tight}
	For any policy $\pi\in\Pi$, under the event $\calE$ we have
	\begin{align*}
		&\max_{\dy\in\caldy}\; \lb(\dy,\pi)\ge\; \eta(\dystar,\pi)-6\vmax\epsilon_\rho(\pi)/(1-\gamma)^2\nonumber\\
		&-\frac{1}{1-\gamma}\Bigl(\vmax\epsilon_\mu(\pi)-2\vmax\sqrt{\zeta\iota/n}-\zeta\vmax\TV{\hmu}{\mu}\Bigr).
	\end{align*}
\end{lemma}
In the sequel, we present a proof sketch for Lemma~\ref{lem:OPE-tight}. In this proof sketch, we hide $1/(1-\gamma)$ factors in the big-O notation. For a fixed policy $\pi$, let $\hdy$ be the minimizer of Eq.~\eqref{equ:error-1}. We prove Lemma~\ref{lem:OPE-tight} by analyzing the terms in the definition of $\lb(\hdy,\pi)$ (Eq.~\eqref{equ:alg-local-lower-bound}) separately. 
\begin{itemize}
	\item[i.] Following the definition of Eq.~\eqref{equ:error-1}, we can show that $\|\rho^\pi_\hdy-\rho^\pi_\dystar\|_1\le \bigO(\epsilon_\rho(\pi)).$ Consequently we get $\eta(\hdy,\pi)\ge \eta(\dystar,\pi)-\bigO(\epsilon_\rho(\pi)).$
	\item [ii.] Recall that $0\le g(s,a)\le \vmax$ for all $g\in\calG$. Then for any $(s,a)$ we have $\sup_{g\in\calG}|\E_{s'\sim\hdy(s,a)}g(s')-\E_{s'\sim\dystar(s,a)}g(s')]|\le \vmax\text{TV}(\hdy(s,a),\dystar(s,a)).$ Combining the definition of $\ell_w(g,\dy)$, Eq.~\eqref{equ:error-1} and statistical error we get 
	$\sup_{g\in \calG}\ell_{w_{\pi,\dy}}(g,\dy)\le \tildeO(\epsilon_\rho(\pi)+1/\sqrt{n}+\vmax\TV{\hmu}{\mu})$ under event $\calE$.
	\item [iii.] For the last term regarding distribution mismatch, we combine Eq.~\eqref{equ:error-2} and Lemma~\ref{lem:helper-TV-to-IS}. We can upper bound this term by $\bigO(\epsilon_\rho(\pi)+\epsilon_\mu(\pi)).$
	\item [iv.] The final term arises due to the potential estimation error in the behavior policy distribution. 
\end{itemize}

Theorem~\ref{thm:OPE-main} follows directly from combining Lemma~\ref{lem:OPE-lowerbound} and Lemma~\ref{lem:OPE-tight}.
Note that Theorem~\ref{thm:OPE-main} accounts for estimation error in the behavior policy, misspecification in the dynamics model class, and misspecification in the value function class, the latter two in a more local, tighter form than prior work. 

\subsection{Illustrative Example}\label{sec:examples}
To build intuition of where our approach may yield benefits, we provide an illustrative example where Alg.~\ref{alg:OPE} has better performance than existing approaches: an MDP whose state space is partitioned into several parts. The model class is restricted so that every model can only be accurate on one part of the state space. When each deterministic policy only visits one part of the state space, the local misspecification error is small --- for each policy, there exists a dynamics model in the set which can accurately estimate the distribution of states and actions visited under that policy. In contrast, if the dynamics are learned to fit the whole state space, the estimation error will be large.

More precisely, for a fixed parameter $d$, consider a  MDP where $\calS=\{s_0,\cdots,s_d\}\cup\{s_{g},s_{b}\}.$ There are $d$ actions %
denoted by $a_1,\cdots,a_d.$ The true dynamics are deterministic and given by  
\begin{align}
	\dystar(s_0,a_i)&=s_i,\label{equ:gtt-1}\quad
	\dystar(s_i,a_j)=\begin{cases}
		s_{g},&\text{if }\ind{i=j},\\
		s_{b},&\text{if }\ind{i\neq j},
	\end{cases}\\
	\dystar(s_g,a_i)&=s_g,
	\quad \dystar(s_b,a_i)=s_b,\forall i\in[d].\label{equ:gtt-4}
\end{align}
And the reward is $r(s,a_i)=\ind{s=s_g},\forall i\in[d]$.

The transition function class $\mathcal{T}$ is parameterized by $\theta\in\R^{d}$. For a fixed $\theta$, the transition for states $s_1,\ldots,s_d$ is 
\begin{align}
	\dy_\theta(s_i,a_j)&=\begin{cases}
		s_{g},&\text{w.p. }\frac{1}{2}\pbra{1+e_j^\top \theta},\\
		s_{b},&\text{w.p. }\frac{1}{2}\pbra{1-e_j^\top \theta},
	\end{cases}
\end{align} where $e_j$ is the $j$-th standard basis of $\R^d$.
The transitions for states $s_0,s_g,s_b$ is identical to the true dynamics $\dystar.$  But the transition model $\dy_\theta$ in the function class must use the same parameter $\theta$ to approximate the dynamics in states  $s_1,\cdots,s_d$, which makes it misspecified.

\paragraph{Decoupling learning the dynamics model and policy is suboptimal.}
Most prior algorithms first learn a dynamics model and then do planning with that model. However, note here that the optimal action induced by MDP planning given a particular $\dy_\theta$ is suboptimal (assuming a uniformly random tie-breaking). This is because, for any given $\theta$, that dynamics model will estimate the dynamics of states $s_1,\cdots,s_d$ as being identical, with identical resulting value functions.  Note this is suboptimality will occur in this example even if the dataset is large and covers the state--action pairs visited by any possible policy ($\epsilon_{\mu}(\pi)=0$), the value function class is tabular and can represent any value function $\epsilon_V = 0$, the behavior policy is known or the resulting estimation error is small ($\TV{\hat{\mu}}{\mu}=0$, and $\zeta=0$). In such a case, Theorem~\ref{thm:OPE-main}  guarantees that with high probability, our algorithm will learn the optimal policy because there exist couplings of the dynamics models and optimal policies such that the local misspecification error $\epsilon_{\rho}=0$. This demonstrates that prior algorithms (including MML \citep{voloshin2021minimax}) that decouple the learning of dynamics and policy can be suboptimal. We now state this more formally:
\begin{theorem}\label{thm:hardinstance}
	Consider any (possibly stochastic) algorithm that outputs an estimated dynamics $\dy_\theta\in \mathcal{T}.$ Let $\pi_\theta$ be the greedy policy w.r.t. $\dy_\theta$ (with ties breaking uniformly at random). Then
	\begin{align}
		\max_{\pi}\eta(T^\star,\pi)-\eta(T^\star,\pi_\theta)\ge \frac{(A-1)\gamma^2}{A(1-\gamma)}.
	\end{align}
\end{theorem}
As a side point, we also show that the off-policy estimation error in \citet{voloshin2021minimax} is large when the dynamics model class is misspecified in Proposition~\ref{prop:loose}. We defer this result to the Appendix. 

\section{Experiments}\label{sec:experiments}
While our primary contribution is theoretical, we now investigate how our method can be used for offline model-based policy selection 
with dynamics model misspecification. 
We first empirically
evaluate our method on Linear-Quadratic Regulator (LQR),
a commonly used environment in optimal control theory \citep{bertsekas2000dynamic}, in order to assess:
\textit{Can Algorithm \ref{alg:OPE} return the optimal policy when we have both model and distribution mismatch?} %
We also evaluate our approach using D4RL~\citep{fu2020d4rl}, a standard offline RL benchmark for continuous control tasks. Here we consider: \textit{Given policies and dynamics pairs obtained using state-of-the-art offline model-based RL methods with ensemble dynamics, does Alg. \ref{alg:OPE} allow picking the best policy, outperforming previous methods?}

\begin{figure*}[htp]
\centering
\begin{subfigure}[b]{0.31\textwidth}
	\includegraphics[width=\linewidth]{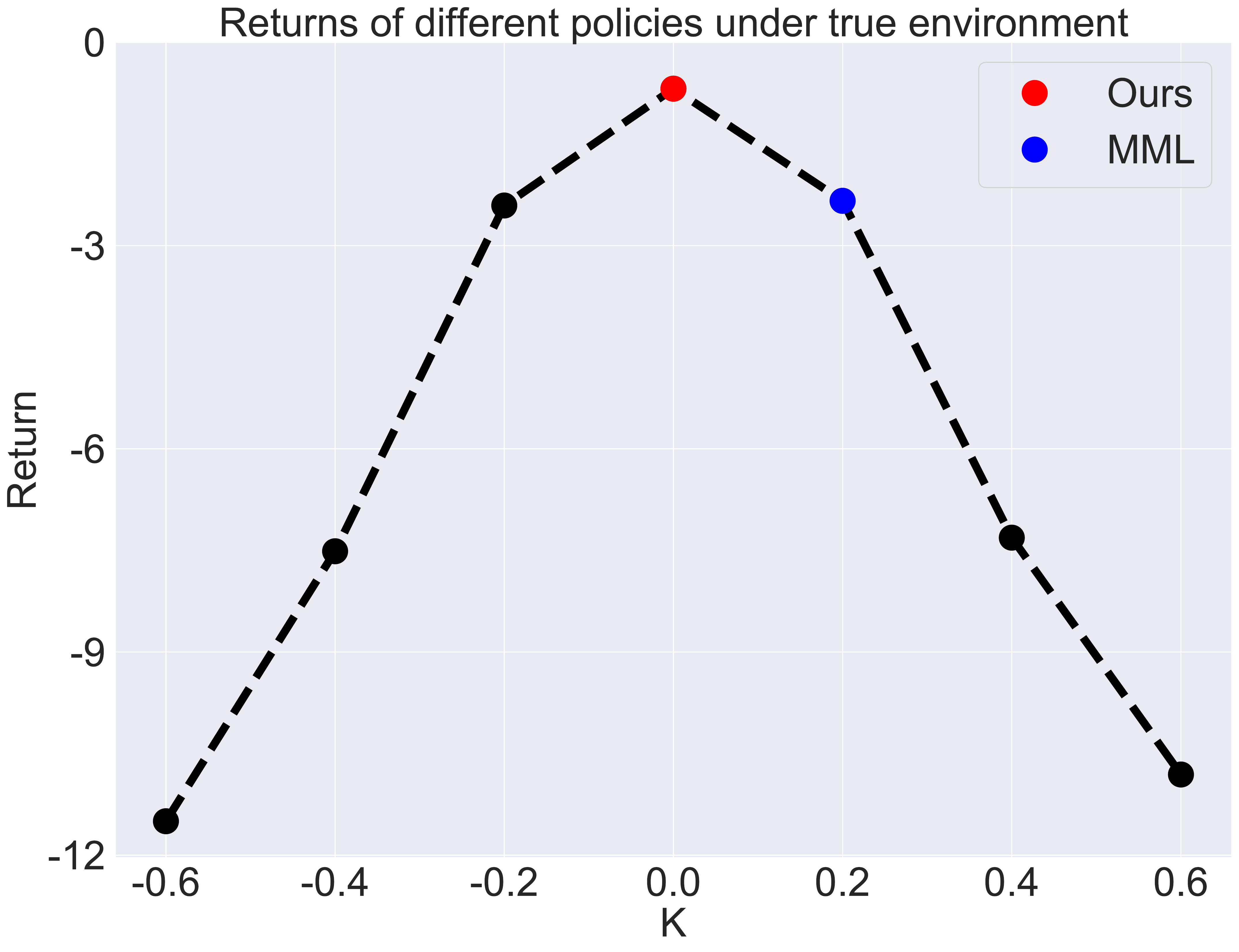}
\end{subfigure}
~~
\begin{subfigure}[b]{0.31\textwidth}
	\includegraphics[width=\linewidth]{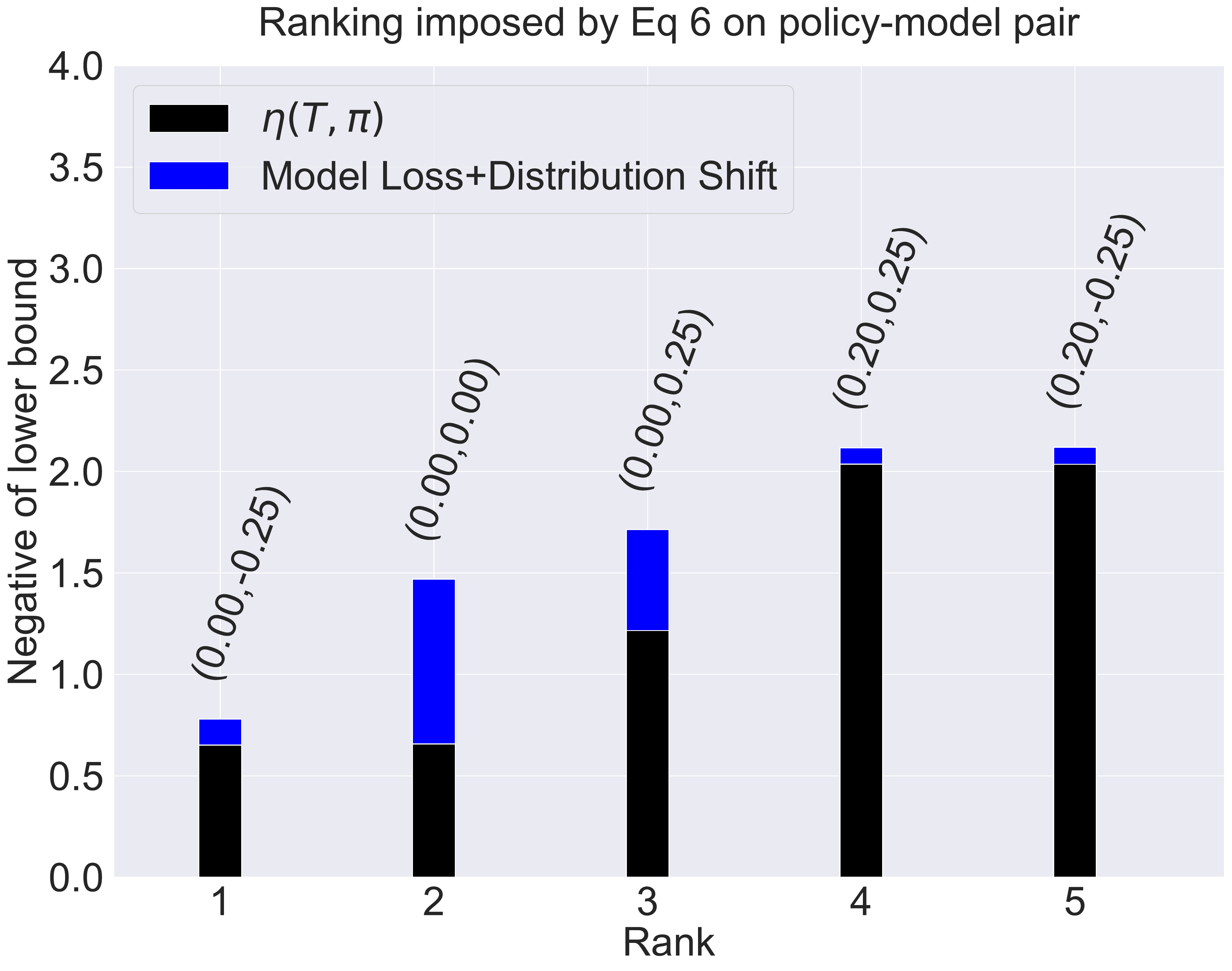}
\end{subfigure}
~~
\begin{subfigure}[b]{0.31\textwidth}
	\includegraphics[width=\linewidth]{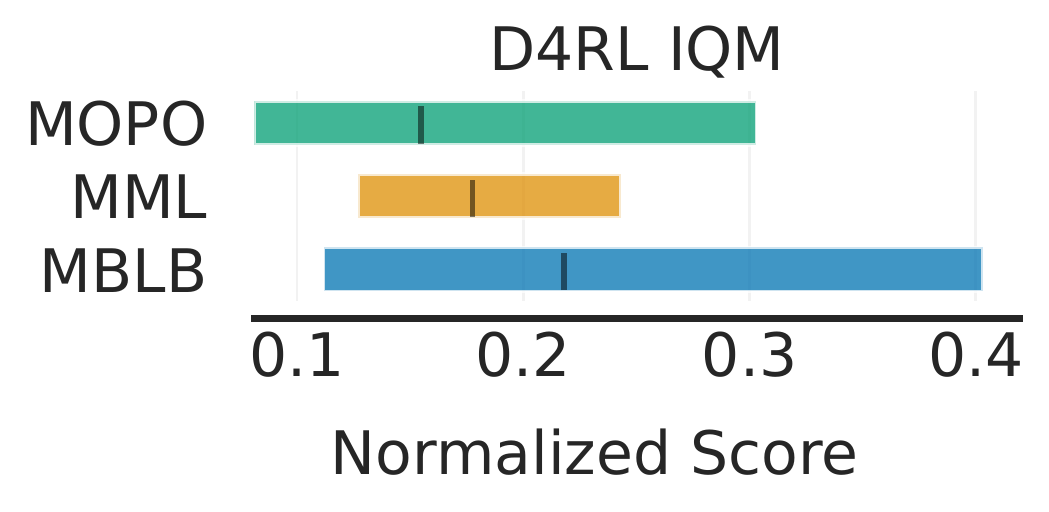}
\end{subfigure}
\caption{\textbf{Left:} Visualization of true policy value $\eta(\dystar,\pi)$. Our algorithm picks the optimal policy, whereas MML picks a suboptimal policy.
\textbf{Middle:} Visualization of negative lower bounds $\lb(\dy,\pi)$ for different policies and models (indexed by the values of $(v,u)$). \textbf{Right:} We show the interquartile mean (IQM) scores of two model-based lower bounds (MML and MBLB) and a recent model-based policy learning algorithm (MOPO) on D4RL.
}
\label{fig:visualization}
\end{figure*}

\subsection{Linear-Quadratic Regulator (LQR)} 

LQR is defined by a linear transition dynamics $s_{t+1} = A s_t + B a_t + \eta$, where $s_t \in \mathbb{R}^n$ and $a_t \in \mathbb{R}^m$ are state and action at time step $t$, respectively. $\eta \sim \mathcal{N}(0,\sigma^2 I)$ is random noise. LQR has a quadratic reward function $\mathcal{R}(s,a) = -(s^T Q s + a^T R a)$ with $Q \in \mathbb{R}^{n\times n}$ and  $R \in \mathbb{R}^{m\times m}$ being positive semi-definite matrices, $Q, R \succeq 0$. The optimal controller to maximize the sum of future rewards $\sum_{t=1}^H -(s_t^TQs_t + a_t^TRa_t)$ until the end of horizon $H$ has the form $a_t = -K s_t$ ($K \in \mathbb{R}^{m\times n}$)~\citep{bertsekas2000dynamic}. The value function is also a quadratic function, $V(s) = s^T U s + q$ for some constant $q$ and positive semi-definite matrix $U \succeq 0$. In the experiment, the state space is $[-1,1].$

\paragraph{Misspecified transition classes.} Consider a $1D$ version of LQR with $A(x) = (1+x/10), B(x)=(-0.5-x/10)$, $Q = 1, R = 1$ and noise $\eta \sim \mathcal{N}(0,0.05)$. Our true dynamics is given by $x^*=6,$ and the corresponding optimal policy has $K=-1.1$. Function classes used by Alg.~\ref{alg:OPE} are finite and computed as follows: (i) the value function class $\calG$ contains the value functions of $1D$ LQR with parameters $x\in \{2,4,10\}$ and $K\in\{-1.1,-0.9,-0.7\}$;
(ii) the transition class $\caldy$ is misspecified. We use the following transition class $\dy_u \in \caldy$ parametrized by $u$,
	\begin{equation*}
		\dy_u=
		\begin{cases}
			s_{t+1} = A(x^*)s_{t} - B(x^*) a_t,&s_t\in [u,u+1],\\
			s_{t+1} = s_t,&\text{otherwise},
		\end{cases}
	\end{equation*}
with $u\in \{-0.75, -0.5, -0.25, 0, 0.25\}.$ In other words, the capacity of the transition class is limited -- each function can only model the true dynamics of a part of the states;
(iii) the policy class is given by $\pi_v$ parameterized by $v$, and
	$\pi_v(s)=-1.1(s-v)+\calN(0,0.01)
	$
	with $v\in \{-0.6, -0.4, -0.2, 0, 0.2, 0.4, 0.6\}.$ Intuitively, $\pi_v$ tries to push the state toward $s=v$.

Since the state and action spaces are one dimensional, we can compute the density ratio $w_{\pi,\dy}$ efficiently by discretization. The implementation details are deferred to Appendix.

\paragraph{Baseline.} We compare our algorithm to 
minimizing MML loss as described in the OPO algorithm of \citet[Algorithm 2]{voloshin2021minimax}. MML
strictly outperformed VAML \cite{farahmand2017value}
as shown in the experiments of \cite{voloshin2021minimax}; hence, we only compare to MML in our experiments.

\paragraph{Results.}
Figure \ref{fig:visualization} (Left) shows the return of different policies under the true environment. %
Our method picks the optimal policy
for the true model, whereas MML picks the wrong policy. %
In Figure~\ref{fig:visualization} (Middle), we also visualize different terms in the definition of $\lb(\dy,\pi)$ (Eq.~\eqref{equ:lowerbound-mis}). Note that the model loss for different policy is different (model loss for $(v,u)=(0,0)$ is significantly larger than $(0.0.-0.25)$, even if the dynamics are the same). This is because the model loss is evaluated with a different density ratio. 

This highlights the main benefit of our method over the baseline. Since the model class is misspecified, maximizing over the weight function $w$ in the MML loss results in an unrealistically large loss value for some models. However, if the chosen policy does not visit the part of the state space with a large error, there is no need to incur a high penalty. %

\begin{table*}[htp]
\centering
\small
\begin{tabular}{@{}lccccccc@{}}
\toprule
Dataset Type & Env & MOPO                                                     & \begin{tabular}[c]{@{}c@{}}MML\\ (Squared)\end{tabular}             & \begin{tabular}[c]{@{}c@{}}MML\\ (Polynomial)\end{tabular} & \multicolumn{1}{l}{\begin{tabular}[c]{@{}c@{}}MML\\ (RKHS)\end{tabular}} & \begin{tabular}[c]{@{}c@{}} \textbf{MBLB}\\ \textbf{(Linear)}\end{tabular} & \begin{tabular}[c]{@{}c@{}} \textbf{MBLB}\\ \textbf{(Quad)} \end{tabular} \\ \midrule
medium       & hopper      & \begin{tabular}[c]{@{}c@{}}175.4 \\ (95.3)\end{tabular}  & \begin{tabular}[c]{@{}c@{}}379.4 \\ (466.4)\end{tabular}            & \begin{tabular}[c]{@{}c@{}}375.6 \\ (459.5)\end{tabular}   & \begin{tabular}[c]{@{}c@{}}375.0 \\ (459.9)\end{tabular}                & \begin{tabular}[c]{@{}c@{}}591.7 \\ (523.1)\end{tabular} &    \begin{tabular}[c]{@{}c@{}}\textbf{808.5} \\ (502.7)\end{tabular}     \\ \midrule
med-expert   & hopper      & \begin{tabular}[c]{@{}c@{}}183.8 \\ (94.4)\end{tabular}  & \begin{tabular}[c]{@{}c@{}}160.9 \\ (131.5)\end{tabular}            & \begin{tabular}[c]{@{}c@{}}116.5 \\ (148.4)\end{tabular}  & \begin{tabular}[c]{@{}c@{}}61.4 \\ (35.0)\end{tabular}                  & \begin{tabular}[c]{@{}c@{}}\textbf{261.1} \\ (157.9)\end{tabular} &   \begin{tabular}[c]{@{}c@{}}242.5 \\ (134.0)\end{tabular}      \\ \midrule
expert       & hopper      & \begin{tabular}[c]{@{}c@{}}80.4 \\ (63.4)\end{tabular}   & \begin{tabular}[c]{@{}c@{}}93.8 \\ (87.9)\end{tabular}              & \begin{tabular}[c]{@{}c@{}}61.6 \\ (61.9)\end{tabular}    & \begin{tabular}[c]{@{}c@{}}70.0 \\ (56.2)\end{tabular}                  & \begin{tabular}[c]{@{}c@{}} 118.2 \\ (61.6)\end{tabular} &     \begin{tabular}[c]{@{}c@{}}\textbf{121.0} \\ (72.5)\end{tabular}   \\ \midrule
medium       & halfcheetah & \begin{tabular}[c]{@{}c@{}}599.8 \\ (668.4)\end{tabular} & \begin{tabular}[c]{@{}c@{}}1967.6 \\ (1707.5)\end{tabular} & \begin{tabular}[c]{@{}c@{}}2625.1 \\ (937.2)\end{tabular} & \textbf{\begin{tabular}[c]{@{}c@{}}3858.2 \\ (1231.1)\end{tabular}}     &  \begin{tabular}[c]{@{}c@{}}3290.4 \\ (1753.1)\end{tabular} &       \begin{tabular}[c]{@{}c@{}}2484.2 \\ (1526.8)\end{tabular}       \\ \midrule
med-expert   & halfcheetah & \begin{tabular}[c]{@{}c@{}}-486.6 \\ (48.1)\end{tabular} & \begin{tabular}[c]{@{}c@{}}-188.5 \\ (137.2)\end{tabular}            & \begin{tabular}[c]{@{}c@{}}-77.0 \\ (252.5)\end{tabular}   & \begin{tabular}[c]{@{}c@{}}-343.2\\ (225.2)\end{tabular}                 & \begin{tabular}[c]{@{}c@{}} \textbf{207.4} \\ (509.5) \end{tabular} &  \begin{tabular}[c]{@{}c@{}} 192.8 \\ (432.0) \end{tabular}   \\
\bottomrule
\end{tabular}
\caption{We report the mean and (standard deviation) of selected policy's simulator environment performance across 5 random seeds. MML and MBLB are used as model-selection procedures where they select the best policy for each seed. Our method is choosing the most near-optimal policy across the datasets.}
\label{tab:d4rl}
\end{table*}

\subsection{D4RL}
D4RL~\citep{fu2020d4rl} is an offline RL standardized benchmark designed and commonly used to evaluate the progress of offline RL algorithms. This benchmark is standard for evaluating offline policy learning algorithms. Here, we use a state-of-the-art policy learning algorithm MOPO~\cite{yu2020mopo} to propose a set of policy-transition model tuples -- for $N$ policy hyperparameters and $K$ transition models, we can get $M \times K$ tuples: $\{(\pi_1, T_1), (\pi_1, T_2), ..., (\pi_N, T_K)\}$. The MOPO algorithm learns an ensemble of transition models and randomly chooses one to sample trajectories during each episode of training. Instead, we choose one transition model to generate trajectories for the policy throughout the entire training. In our experiment, we choose $M=1$ and $K=5$, and train each tuple for 5 random seeds on Hopper and HalfCheetah tasks (see Appendix). We then compute the model-based lower bound for each $(\pi_i, T_j)$, and select the optimal policy that has the highest lower bound.
We learn the dynamics using 300k iterations and we train each policy using 100k gradient iterations steps with SAC~\citep{haarnoja2018} as the policy gradient algorithm, imitating MOPO~\citep{yu2020mopo} policy gradient update.

\subsubsection{MML.}

\citet{voloshin2021minimax} recommended two practical implementations for computing MML lower bounds. The implementation parametrizes $w(s, a) V(s')$ jointly via a new function $h(s, a, s')$. We refer readers to Prop 3.5 from \citet{voloshin2021minimax} for a detailed explanation. We describe how we parametrize this function as follows:

\begin{itemize}[noitemsep,topsep=0pt]
    \item \textbf{Linear}: \citet{voloshin2021minimax} showed that if $T, V, \mu$ are all from the linear function classes, then a model $T$ that minimizes MML loss is both unique and identifiable. This provides a linear parametrization of $h(s, a, s') = \psi(s, a, s')^T \theta$, where $\psi$ is a basis function. We choose $\psi$ to be either a \textbf{\textit{squared}} basis function or a \textbf{\textit{polynomial}} basis function with degree 2. 
    \item \textbf{Kernel}: Using a radial basis function (RBF) over $\mathcal{S} \times \mathcal{A} \times \mathcal{S}$ and computing $K((s, a, s'), (\tilde s, \tilde a, \tilde s'))$, \citet{voloshin2021minimax} showed that there exists a closed-form solution to compute the maxima of the MML loss (\textbf{RKHS}). Here, there is no need for any gradient update, we only sample $s'$ from $T$.
\end{itemize}

\subsubsection{MBLB (Ours).}

For a continuous control task, we compute our  model-based lower bound (MBLB) as follows:

\paragraph{Compute $\eta(T, \pi)$.} 
Although it is reasonable to directly use a value function $V^{\pi}_T$ trained during policy learning to compute $\eta(T, \pi)$, \citet{paine2020hyperparameter,kumar2021a} points out how this value function often severely over-estimates the actual discounted return. Therefore, 
 we estimate the expected value of policy $\pi$ using the generalized advantage estimator (GAE)~\citep{schulman2015high}. For a sequence of transitions $\{s_t, a_t, r(s_{t},a_{t}), s_{t + 1}\}_{t \in [0, N]}$, it is defined as: $A_t = \sum_{t'=t}^{t + N} (\gamma \lambda)^{t' - t} (r(s_{t'},a_{t'}) + \gamma V_{\phi}(s_{t' + 1}) - V_{\phi}(s_{t'}))$, with $\lambda$ a fixed hyperparameter and $V_{\phi}$ the value function estimator at the previous optimization iteration. Then, to estimate the value function, we solve the non-linear regression problem $\operatorname{minimize}_{\phi}\sum_{t'=t}^{t+N}(V_{\phi}(s_{t'})-\hat{V}_{t'})^{2}$ where $\hat{V}_{t} = A_t + V_{\phi}(s_{t'})$. We also provide a comparison to using the standard TD-1 Fitted Q Evaluation (FQE)~\citep{le2019batch} instead in Table~\ref{tab:fqe_d4rl} in the Appendix. We find that using GAE provides better policy evaluation estimations.

\paragraph{Behavior density modeling.} We use a state-of-the-art normalizing flow probability model to estimate the density of state-action pairs~\cite{papamakarios2021normalizing}. For $\rho^{\pi}_T$, we sample 10,000 trajectories from $T, \pi$, and estimate the corresponding density; for the behavior distribution $\mu$, we use the given dataset $\calD$. We empirically decide the number of training epochs that will give the model the best fit.

\paragraph{Compute $\sup_{g \in \mathcal{G}} |\ell_{w_{\pi, T}}(g, T)|$.} We parametrize $g$ either as a linear function of state: $g(s) = m^T s$, or a quadratic function of the state: $g(s) = s^T M s + b$. We use gradient ascent on $\ell_{w_{\pi, T}}(g, T)$ to  maximize this objective.

\paragraph{Results.}
We report the results in Table~\ref{tab:d4rl}. There is general overlap across seeds for the performance between various methods, but our approach has the best average performance or is within the standard deviation of the best. We also show that for different choices of how we parameterize the $w(s, a)V(s')$ distribution (MML) and how we choose the family of $g$ test function (MBLB), we are selecting different final policies. However, overall, MBLB can pick better-performing final policies with two different parametrizations while MML is choosing lower-performing policies with its three parametrizations. We find that our approach of selecting among the set of policies computed from each of the models used by MOPO consistently outperforms the policy produced by MOPO in the considered tasks.

To summarize these results, 
we report the interquartile mean (IQM) scores of each method in Figure~\ref{fig:visualization} (Right). IQM is an outlier robust metric proposed by \citet{agarwal2021deep} to compare deep RL algorithms. We create the plot by sampling with replacement over all runs on all datasets 50000 times. Though there is significant overlap, our method generally outperforms policies learned from MOPO.

\section{Conclusion}\label{sec:conclusion}
There are many directions for future work. The current  $\lb(\dy,\pi)$ implementation with density ratio $w_{\pi,\dy}(s,a)$ is not differentiable: an interesting question is to make this differentiable so that we can directly optimize a policy. Another interesting question would be to construct estimators for the local misspecification errors $\epsilon_\rho,\epsilon_\mu$ and $\epsilon_V$, which could be used to refine the model class to optimize performance.

 To conclude, this paper studies model-based offline reinforcement learning with local model misspecification errors, and proves a novel safe policy improvement theorem. Our theoretical analysis shows the benefit of this tighter analysis and approach. We illustrate the advantage of our method over prior work in a small linear quadratic example and also demonstrate that it is competitive or has stronger performance than recent model-based offline RL methods on policy selection in a set of D4RL tasks.

\section*{Acknowledgment}
Research reported in this paper was sponsored in part by NSF grant \#2112926, the DEVCOM Army Research Laboratory under Cooperative Agreement W911NF-17-2-0196 (ARL IoBT CRA) and a Stanford Hoffman-Yee grant. The views and conclusions contained in this document are those of the authors and should not be interpreted as representing the official policies, either expressed or implied, of the Army Research Laboratory or the U.S.Government. The U.S. Government is authorized to reproduce and distribute reprints for Government purposes notwithstanding any copyright notation herein.

\bibliography{aaai22.bib}

\newpage
\setcounter{figure}{0}
\renewcommand{\thefigure}{A\arabic{figure}}
\setcounter{table}{0}  %
\renewcommand{\thetable}{A\arabic{table}}

\appendix
\onecolumn
\section{Missing Proofs}\label{app:proofs}

\subsection{High Probability Events}\label{app:event}
In this section, we introduce concentration inequalities and define the high probability events. 

Define the following quantities
\begin{align}
	&L(\pi,g,\dy)=\E_{(s,a,s')\sim \mu}\bbra{w_{\pi,\dy}(s,a)(\E_{x\sim \dy(s,a)}[g(x)]-\E_{x\sim \dystar(s,a)}[g(x)])},\\
	&l(\pi,g,\dy)=\E_{(s,a,s')\sim \calD}\bbra{w_{\pi,\dy}(s,a)(f^g_\dy(s,a)-g(s'))}.
\end{align}
Recall that $\iota=\log(2|\calG||\caldy||\Pi|/\delta).$ Consider the event 
\begin{align}\label{equ:event}
	\calE=\left\{\abs{L(\pi,g,\dy)-l(\pi,g,\dy)}\le 2\vmax\sqrt{\frac{\zeta\iota}{n}},\quad \forall \pi\in\Pi,g\in\calG,\dy\in\caldy\right\}.
\end{align}
In the following, we show that 
\begin{align}\label{equ:finite-sample-main}
	\Pr\pbra{\calE}\ge 1-\delta.
\end{align}

Recall that $\calD=\{(s_i,a_i,s'_i)\}_{i=1}^{n}$ where $(s_i,a_i,s'_i)\sim \mu$ are i.i.d. samples from distribution $\mu$. For fixed $\pi\in\Pi,g\in\calG,\dy\in\caldy$, we have
$\E[\hat{l}(\pi,g,\dy)]=l(\pi,g,\dy).$ Meanwhile, note that 
\begin{align}
	&\abs{w_{\pi,\dy}(s,a)(f^g_\dy(s,a)-g(s'))}\le \zeta\vmax,\\
	&\E_{(s,a,s')\sim \mu}[w_{\pi,\dy}(s,a)^2(f^g_\dy(s,a)-g(s'))^2]\\
	&\quad\le \E_{(s,a,s')\sim \rho^\pi_\dy}[w_{\pi,\dy}(s,a)(f^g_\dy(s,a)-g(s'))^2]\le\vmax^2\zeta.
\end{align}
By Bernstein inequality, with probability at least $1-\delta/(|\calG||\caldy||\Pi|)$, 
\begin{align}
	\abs{L(\pi,g,\dy)-l(\pi,g,\dy)}\le \sqrt{\frac{2\vmax^2\zeta\log(2|\calG||\caldy||\Pi|/\delta)}{n}}+\frac{\zeta\vmax}{3n}\log(2|\calG||\caldy||\Pi|/\delta)
\end{align}
Recall that $\iota=\log(2|\calG||\caldy||\Pi|/\delta)$. When $n\ge \zeta$ we have
\begin{align}
	\abs{L(\pi,g,\dy)-l(\pi,g,\dy)}\le 2\vmax\sqrt{\frac{\zeta\iota}{n}}.
\end{align}
Note that when $n<\zeta$, $\calE$ trivially holds. As a result, applying union bound we prove Eq.~\eqref{equ:finite-sample-main}.

\subsection{Proof of Lemma~\ref{lem:OPE-lowerbound}}\label{app:pf-OPE-lowerbound}
\begin{proof}
	In the following, we consider a fixed policy $\pi$ and dynamics $\dy\in \caldy$. We use $w$ to denote $w_{\pi,\dy}$ when the context is clear.

	By basic algebra we get
	\begin{align}
		&\abs{\E_{(s,a)\sim \rho^\pi_\dy}\bbra{G^\pi_\dy(s,a)}}\\
		\le\;&\abs{\E_{(s,a)\sim \rho^\pi_\dy}\bbra{\ind{\frac{\rho^\pi_\dy(s,a)}{\hmu(s,a)}\le \zeta}G^\pi_\dy(s,a)}}+\E_{(s,a)\sim \rho^\pi_\dy}\bbra{\ind{\frac{\rho^\pi_\dy(s,a)}{\hmu(s,a)}> \zeta}\abs{G^\pi_\dy(s,a)}}\\
		\le\;&\abs{\E_{(s,a)\sim \hmu}\bbra{w(s,a)G^\pi_\dy(s,a)}}+\vmax\E_{(s,a)\sim \rho^\pi_\dy}\bbra{\ind{\frac{\rho^\pi_\dy(s,a)}{\hmu(s,a)}> \zeta}}.\label{equ:pf-lem43-0}
	\end{align}
	Note that 
	\begin{align}
	    \E_{ (s,a) \sim \hat{\mu} } [ w(s,a) G^\pi_T (s,a) ] &= \sum_{s,a} \hat{\mu}(s,a) w(s,a) G^\pi_T (s,a)\\
&= \sum_{s,a} ( \hat{\mu}(s,a) - \mu(s,a) + \mu(s,a) ) w(s,a) G^\pi_T (s,a)\\
&= \sum_{s,a} \mu(s,a)   w(s,a) G^\pi_T (s,a) + \sum_{s,a} (\hat{\mu}(s,a) - \mu(s,a)) w(s,a) G^\pi_T (s,a)\\
&\le \E_{ (s,a) \sim \mu } [ w(s,a) G^\pi_T (s,a) ] + \sum_{s,a} |\hat{\mu}(s,a) - \mu(s,a)| \zeta V_{\rm max}\\
&\le \E_{ (s,a) \sim \mu } [ w(s,a) G^\pi_T (s,a) ] + \zeta V_{\rm max} \TV{\hat\mu}{\mu}.
	\end{align}
	Continuing Eq.~\eqref{equ:pf-lem43-0} we get
	\begin{align}
		&\abs{\E_{(s,a)\sim \rho^\pi_\dy}\bbra{G^\pi_\dy(s,a)}}\\\le\;&\abs{\E_{(s,a)\sim \mu}\bbra{w(s,a)G^\pi_\dy(s,a)}}+\vmax\E_{(s,a)\sim \rho^\pi_\dy}\bbra{\ind{\frac{\rho^\pi_\dy(s,a)}{\hmu(s,a)}> \zeta}}+\zeta\vmax\TV{\hmu}{\mu}.\label{equ:pf-lem43-1}
	\end{align}
	
	Consequently, in the following we prove
	\begin{align*}
		&\abs{\E_{(s,a)\sim \mu}\bbra{w(s,a)G^\pi_\dy(s,a)}}
		\le \sup_{g\in \calG}\ell_w(g,\dy)
		 +\epsilon_V(\dy,\pi)+2\vmax\sqrt{\frac{\zeta\iota}{n}}.
	\end{align*}
	Let $L_w(g,\dy)=\abs{\E_{(s,a,s')\sim \mu}\bbra{w(s,a)(\E_{x\sim \dy(s,a)}[g(x)]-\E_{x\sim \dystar(s,a)}[g(x)])}}$ be the population error. Recall that under the high probability event $\calE$ in Eq.~\eqref{equ:event}, for any $g\in\calG$ and $\dy\in\caldy$
	\begin{align}\label{equ:pf-finite-sample}
		\abs{L_w(g,\dy)-\ell_w(g,\dy)}\le 2\vmax\sqrt{\frac{\zeta\iota}{n}}.
	\end{align}
	
	Now by the definition of $G_\dy^\pi(s,a)$, for any $g\in\calG$ we have
	\begin{align}
		&\abs{\E_{(s,a)\sim \mu}\bbra{w(s,a)G^\pi_\dy(s,a)}}\\
		=\;&\abs{\E_{(s,a)\sim \mu}\bbra{w(s,a)\pbra{\E_{s'\sim \dy(s,a)}[\valstar(s')]-\E_{s'\sim \dystar(s,a)}[\valstar(s')]}}}\\
		\le\; &\abs{\E_{(s,a)\sim \mu}\bbra{w(s,a)\pbra{\E_{s'\sim \dy(s,a)}[g(s')]-\E_{s'\sim \dystar(s,a)}[g(s')]}}}\\
		& +\abs{\E_{(s,a)\sim \mu}\bbra{w(s,a)\pbra{\E_{s'\sim \dy(s,a)}\bbra{g(s')-V^\pi_\dystar(s')}+\E_{s'\sim \dystar(s,a)}\bbra{g(s')-V^\pi_\dystar(s')}}}}.\label{equ:lb-4}
	\end{align}
	Define $$\hat{g}=\argmin_{g\in \calG}\abs{\E_{(s,a)\sim \mu}\left[w(s,a)\left(\E_{s'\sim \dy(s,a)}\bbra{g(s')-V^\pi_\dystar(s')}
	+\E_{s'\sim \dystar(s,a)}\bbra{g(s')-V^\pi_\dystar(s')}\right)\right]}.$$
	Since $g$ is arbitrarily, continuing Eq.~\eqref{equ:lb-4} and recalling Definition~\ref{def:local-mis} we get
	\begin{align}
		&\abs{\E_{(s,a)\sim \mu}\bbra{w(s,a)G^\pi_\dy(s,a)}}\\
		\le\; &\abs{\E_{(s,a)\sim \mu}\bbra{w(s,a)\pbra{\E_{s'\sim \dy(s,a)}[\hat{g}(s')]-\E_{s'\sim \dystar(s,a)}[\hat{g}(s')]}}}+\epsilon_V(T,\pi)\\
		\le\; &\sup_{g\in\calG}\abs{\E_{(s,a)\sim \mu}\bbra{w(s,a)\pbra{\E_{s'\sim \dy(s,a)}[g(s')]-\E_{s'\sim \dystar(s,a)}[g(s')]}}}+\epsilon_V(T,\pi).
		\label{equ:lb-5}
	\end{align}
	Combining Eq.~\eqref{equ:lb-5} and Eq.~\eqref{equ:pf-finite-sample} we get,
	\begin{align}
		&\abs{\E_{(s,a)\sim \mu}\bbra{w(s,a)G^\pi_\dy(s,a)}}
		\le \sup_{g\in \calG}L_w(g,\dy)+\epsilon_V(T,\pi)\\
		\le\;& \sup_{g\in \calG}\ell_w(g,\dy)+\epsilon_V(T,\pi)+2\vmax\sqrt{\frac{\zeta\iota}{n}}.
	\end{align}
	Now plugging in Eq.~\eqref{equ:pf-lem43-1} we get,
	\begin{align*}
		&\abs{\E_{(s,a)\sim \rho^\pi_\dy}\bbra{G^\pi_\dy(s,a)}}\\
		\le\;& \sup_{g\in \calG}\ell_w(g,\dy)+\epsilon_V(T,\pi)+2\vmax\sqrt{\frac{\zeta\iota}{n}}
		+\vmax\E_{(s,a)\sim \rho^\pi_\dy}\bbra{\ind{\frac{\rho^\pi_\dy(s,a)}{\hmu(s,a)}> \zeta}}+\zeta\vmax\TV{\hmu}{\mu}.
	\end{align*}
	Finally, combining with simulation lemma (Lemma~\ref{lem:simulation}) we finish the proof.
\end{proof}

\subsection{Proof of Lemma~\ref{lem:OPE-tight}}\label{app:pf-OPE-tight}

\begin{proof}[Proof of Lemma~\ref{lem:OPE-tight}]
	Consider a fixed $\pi\in\Pi$. When the context is clear, we use $\epsilon_\rho$ and $\epsilon_\mu$ to denote $\epsilon_\rho(\pi)$ and $\epsilon_\mu(\pi)$ respectively.
	
	Consider the dynamics 
	\begin{align}
		\hdy=\argmin_{\dy\in\caldy}\E_{(s,a)\sim \rho^\pi_\dystar}[\TV{\dy(s,a)}{\dystar(s,a)}].
	\end{align}
	By the definition of $\epsilon_\rho$ we get $$\E_{(s,a)\sim \rho^\pi_\dystar}\bbra{\TV{\hdy(s,a)}{\dystar(s,a)}}\le \epsilon_\rho.$$
	Applying Lemma~\ref{lem:TV-chainrule} we get
	\begin{equation}\label{equ:pf-local-1}
		\normone{\rho^\pi_\hdy-\rho^\pi_\dystar}\le \frac{\epsilon_\rho}{(1-\gamma)}.
	\end{equation} 

	The rest of the proof is organized in the following way. We bound the three terms in RHS of Eq.~\eqref{equ:alg-local-lower-bound} respectively as follows
	\begin{align}
		&\eta(\hdy,\pi)\ge \eta(\dystar,\pi)-\frac{\vmax}{1-\gamma}\epsilon_\rho,\label{equ:pf-local-2}\\
		&\sup_{g\in \calG}\ell_w(g,\hdy)\le \frac{2\vmax\epsilon_\rho}{1-\gamma}+2\vmax\sqrt{\frac{\zeta\iota}{n}}+\zeta\vmax\TV{\hmu}{\mu},\label{equ:pf-local-3}\\
		&\E_{(s,a)\sim\rho^\pi_\hdy}\bbra{\ind{\frac{\rho^\pi_\hdy(s,a)}{\hmu(s,a)}> \zeta}}\le \epsilon_\mu+\frac{3\epsilon_\rho}{(1-\gamma)}.\label{equ:pf-local-4}
	\end{align}
	Then we combine these inequalities together to prove Lemma~\ref{lem:OPE-tight}.
	
	\paragraph{Step 1: Proving Eq.~\eqref{equ:pf-local-2}.} Note that for every $\dy$ and $\pi$, $\eta(\dy,\pi)=\frac{1}{1-\gamma}\dotp{\rho^\pi_\dy}{r}$ where $r$ is the reward function. Then we have \begin{equation}
		\eta(\dystar,\pi)-\eta(\hdy,\pi)=\frac{1}{1-\gamma}\dotp{\rho^\pi_\dystar-\rho^\pi_\hdy}{r}\le \frac{1}{1-\gamma}\normone{\rho^\pi_\dystar-\rho^\pi_\hdy}\norm{r}_\infty.
	\end{equation}
	Combining with Eq.~\eqref{equ:pf-local-1} we get Eq.~\eqref{equ:pf-local-2}.
	
	\paragraph{Step 2: Proving Eq.~\eqref{equ:pf-local-3}.} For any fixed function $g\in\calG$.
	Let $w=w_{\pi,\hdy}$ be a shorthand. Define $$L_w(g,\dy)=\abs{\E_{(s,a,s')\sim \mu}\bbra{w(s,a)(f^g_\dy(s,a)-g(s'))}}$$ to be the population error. 
	Then we have
	\begin{align*}
		&L_w(g,\hdy)\\
		=\;& \abs{\E_{(s,a)\sim \mu}\bbra{w(s,a)\pbra{\E_{s'\sim \hdy(s,a)}[g(s')]-\E_{s'\sim \dystar(s,a)}[g(s')]}}}\\
		\le\;& \abs{\E_{(s,a)\sim \hmu}\bbra{w(s,a)\pbra{\E_{s'\sim \hdy(s,a)}[g(s')]-\E_{s'\sim \dystar(s,a)}[g(s')]}}}+\zeta\vmax\TV{\hmu}{\mu}\\
		=\;& \abs{\E_{(s,a)\sim \rho^{\pi}_\hdy}\bbra{\ind{\frac{\rho^\pi_\hdy(s,a)}{\hmu(s,a)}\le \zeta}\pbra{\E_{s'\sim \hdy(s,a)}[g(s')]-\E_{s'\sim \dystar(s,a)}[g(s')]}}}+\zeta\vmax\TV{\hmu}{\mu}\\
		\le\;& \vmax\E_{(s,a)\sim \rho^{\pi}_\hdy}\bbra{\ind{\frac{\rho^\pi_\hdy(s,a)}{\hmu(s,a)}\le \zeta}\TV{\hdy(s,a)}{\dystar(s,a)}}+\zeta\vmax\TV{\hmu}{\mu}\\
		\le\;& \vmax\E_{(s,a)\sim \rho^{\pi}_\dystar}\bbra{\TV{\hdy(s,a)}{\dystar(s,a)}}+\frac{\vmax\epsilon_\rho}{1-\gamma}\tag{By Eq.~\eqref{equ:pf-local-1}}+\zeta\vmax\TV{\hmu}{\mu}\\
		\le\;& \vmax\pbra{\epsilon_\rho+\frac{\epsilon_\rho}{1-\gamma}}+\zeta\vmax\TV{\hmu}{\mu}\le \frac{2\vmax\epsilon_\rho}{1-\gamma}+\zeta\vmax\TV{\hmu}{\mu}.
	\end{align*}
	Under event $\calE$ we have 
	\begin{align}
		\ell_w(g,\hdy)\le L_w(g,\hdy)+2\vmax\sqrt{\frac{\zeta\iota}{n}}.
	\end{align}
	Because $g$ is arbitrary, we get Eq.~\eqref{equ:pf-local-3}.
	
	\paragraph{Step 3: Proving Eq.~\eqref{equ:pf-local-4}.} Note that
	\begin{align}
		&\E_{(s,a)\sim\rho^\pi_\hdy}\bbra{\ind{\frac{\rho^\hpi_\dy(s,a)}{\hmu(s,a)}> \zeta}}\\
		=\;&\E_{(s,a)\sim\rho^\pi_\hdy}\bbra{\ind{\frac{\rho^\pi_\hdy(s,a)}{\rho^\pi_\dystar(s,a)}\frac{\rho^\pi_\dystar(s,a)}{\hmu(s,a)}> \zeta}}\\
		\le\;&\E_{(s,a)\sim\rho^\pi_\hdy}\bbra{\ind{\frac{\rho^\pi_\hdy(s,a)}{\rho^\pi_\dystar(s,a)}>2}}+\E_{(s,a)\sim\rho^\pi_\hdy}\bbra{\ind{\frac{\rho^\pi_\dystar(s,a)}{\hmu(s,a)}> \zeta/2}}.\label{equ:pf-step4-1}
	\end{align}
	With the help of Lemma~\ref{lem:helper-TV-to-IS}, we can upper bound the first term of Eq.~\eqref{equ:pf-step4-1} by the total variation between $\rho^\pi_\hdy$ and $\rho^\pi_\dystar.$ Combining Lemma~\ref{lem:helper-TV-to-IS} and Eq.~\eqref{equ:pf-local-1} we get
	\begin{equation}
		\E_{(s,a)\sim\rho^\pi_\hdy}\bbra{\ind{\frac{\rho^\hpi_\dy(s,a)}{\rho^\pi_\dystar(s,a)}>2}}\le \frac{2\epsilon_\rho}{1-\gamma}.
	\end{equation}
	On the other hand, by combining Eq.~\eqref{equ:pf-local-1} and the definition of $\epsilon_\mu$ we get
	\begin{equation*}
		\E_{(s,a)\sim\rho^\pi_\hdy}\bbra{\ind{\frac{\rho^\pi_\dystar(s,a)}{\hmu(s,a)}> \zeta/2}}\le \E_{(s,a)\sim\rho^\pi_\dystar}\bbra{\ind{\frac{\rho^\pi_\dystar(s,a)}{\hmu(s,a)}> \zeta/2}}+\frac{\epsilon_\rho}{1-\gamma}\le \epsilon_\mu+\frac{\epsilon_\rho}{1-\gamma}.
	\end{equation*}
	Consequently, we get Eq.~\eqref{equ:pf-local-4}.
	
	Now we stitch Eq.~\eqref{equ:pf-local-1}, Eq.~\eqref{equ:pf-local-2} and Eq.~\eqref{equ:pf-local-3} together. Combining with the definition of $\lb(\hdy,\pi)$ in Eq.~\eqref{equ:alg-local-lower-bound}, we have
	\begin{align*}
		\lb(\hdy,\pi)=\;&\eta(\hdy,\pi)-\frac{1}{1-\gamma}\pbra{\sup_{g\in \calG}\abs{\ell_{w_{\pi,\dy}}(g,\hdy)}+\vmax\E_{(s,a)\sim\rho^\pi_\dy}\bbra{\ind{\frac{\rho^\pi_\hdy(s,a)}{\hmu(s,a)}> \zeta}}+2\zeta\vmax\TV{\hmu}{\mu}}\\
		\ge\;& \eta(\dystar,\pi)-\frac{\vmax\epsilon_\rho}{1-\gamma}-\frac{2\vmax\epsilon_\rho}{(1-\gamma)^2}-\frac{2\vmax}{1-\gamma}\sqrt{\frac{\zeta\iota}{n}}-\frac{\vmax}{1-\gamma}\pbra{\frac{3\epsilon_\rho}{1-\gamma}+\epsilon_\mu}-\frac{2\zeta\vmax\TV{\hmu}{\mu}}{1-\gamma}\\
		\ge\;&\eta(\dystar,\pi)-\frac{6\vmax\epsilon_\rho}{(1-\gamma)^2}-\frac{\vmax\epsilon_\mu}{1-\gamma}-\frac{2\vmax}{1-\gamma}\sqrt{\frac{\zeta\iota}{n}}-\frac{2\zeta\vmax\TV{\hmu}{\mu}}{1-\gamma}.
	\end{align*}
	Note that $\hdy\in\caldy$, we have
	\begin{align}
		\max_{\dy\in\caldy}\lb(\dy,\pi)\ge \lb(\hdy,\pi),
	\end{align}
	which finishes the proof.
\end{proof}

\subsection{Proof of Theorem~\ref{thm:OPE-main}}\label{app:pf-OPE-main}
\begin{proof}[Proof of Theorem~\ref{thm:OPE-main}]
	Let $\hdy,\hpi\gets \argmax_{\dy\in\caldy,\pi\in\Pi} \lb(\dy,\pi)$ be the dynamics and policy that maximizes the lower bound. Note that $\hpi$ is the output of Algorithm~\ref{alg:OPE}.
	
	Now under the event $\calE$, by Lemma~\ref{lem:OPE-tight}, for any policy $\pi$ we have
	\begin{align}\label{equ:thm-main-1}
		\max_{\dy\in\caldy} \lb(\dy,\pi)\ge \eta(\dystar,\pi)-\frac{6\vmax\epsilon_\rho(\pi)}{(1-\gamma)^2}-\frac{\vmax\epsilon_\mu(\pi)}{1-\gamma}-\frac{2\vmax}{1-\gamma}\sqrt{\frac{\zeta\iota}{n}}-\frac{2\zeta\vmax\TV{\hmu}{\mu}}{1-\gamma}.
	\end{align}
	On the other hand, under the event $\calE$, by Lemma~\ref{lem:OPE-lowerbound} we get
	\begin{align}\label{equ:thm-main-2}
		\eta(\dy^\star,\pi)
		\ge \lb(\hdy,\hpi)-\frac{\epsilon_V(\hdy,\hpi)}{1-\gamma}-\frac{2\vmax}{1-\gamma}\sqrt{\frac{\zeta\iota}{n}}.
	\end{align}
	By the optimality of $\hdy,\hpi$, we have $\lb(\hdy,\hpi)\ge \sup_{\dy\in\caldy}\lb(\dy,\pi)$ for any $\pi$. As a result, combining with Eq.~\eqref{equ:thm-main-1} and Eq.~\eqref{equ:thm-main-2} we get
	\begin{align}
		\eta(\dystar,\hpi)&\ge \lb(\hdy,\hpi)-\frac{\epsilon_V(\hdy,\hpi)}{1-\gamma}-\frac{2\vmax}{1-\gamma}\sqrt{\frac{\zeta\iota}{n}}\\
		&\ge \sup_{\pi\in\Pi}\sup_{\dy\in\caldy}\lb(\dy,\pi)-\frac{\epsilon_V(\hdy,\hpi)}{1-\gamma}-\frac{2\vmax}{1-\gamma}\sqrt{\frac{\zeta\iota}{n}}\\
		&\ge \sup_{\pi}\left\{\eta(\dystar,\pi)-\frac{6\vmax\epsilon_\rho(\pi)}{(1-\gamma)^2}-\frac{\vmax\epsilon_\mu(\pi)}{1-\gamma}\right\}-\frac{\epsilon_V(\hdy,\hpi)}{1-\gamma}-\frac{4\vmax}{1-\gamma}\sqrt{\frac{\zeta\iota}{n}}-\frac{2\zeta\vmax\TV{\hmu}{\mu}}{1-\gamma}.
	\end{align}
\end{proof}

\subsection{Proof of Theorem~\ref{thm:hardinstance}}\label{app:pf-hardinstance}
\begin{proof}[Proof of Theorem~\ref{thm:hardinstance}]
	Note that for any fixed $\theta\in\R^{d}$, the transition function for state $s_1,\cdots,s_d$ are identical. As a result, $Q^\pi_{T_\theta}(s_i,a_j)=Q^\pi_{T_\theta}(s_{i'},a_j),\forall i,i'\in[d]$ for any policy $\pi$.
	Recall that $\pi_\theta$ is the optimal policy of $T_\theta$ (with ties breaking uniformly at random). Therefore, $\pi_{\theta}(s_0)=1/A$ and $\pi_{\theta}(s_i)=\pi_{\theta}(s_{i'}),\forall i,i'\in[d].$
	
	By the definition of the ground-truth dynamics $T^\star$ in Eqs.~\eqref{equ:gtt-1}-\eqref{equ:gtt-4}, we have
	$Q^{\pi_\theta}_{T^\star}(s_i,a_j)=\ind{i=j}\frac{\gamma}{1-\gamma}.$ Therefore,
	\begin{align}
	    \eta(T^\star,\pi_\theta)=\frac{\gamma}{A}\sum_{i=1}^{d}Q^{\pi_\theta}_{T^\star}(s_i,\pi_\theta(s_i))\le \frac{\gamma}{A}\max_a\sum_{i=1}^{d}Q^{\pi_\theta}_{T^\star}(s_i,a)\le \frac{\gamma^2}{A(1-\gamma)}.
	\end{align}
	Since $\max_\pi \eta(T^\star,\pi)=\frac{\gamma^2}{1-\gamma}$, we have $$\max_\pi \eta(T^\star,\pi)-\eta(T^\star,\pi_\theta)\ge \frac{(A-1)\gamma^2}{A(1-\gamma)}.$$
\end{proof}

\subsection{OPE Error of MML}\label{app:pf-MML}
In this section, we show that the off-policy estimation error in \citet{voloshin2021minimax} can be large when the dynamics model class is misspecified in Proposition~\ref{prop:loose}.

The MML algorithm requires an density ratio class $\calW:\calS\times\calA\to\R_+$ and prove that when $w_{\pi,T}\in\calW$ and $V^\pi_\dystar\in\calG$,
\begin{align}\label{equ:mml-main}
	\abs{\eta(\dy,\pi)-\eta(\dystar,\pi)}\le \gamma \min_{\dy\in\caldy}\max_{w\in\calW,g\in\calG}\abs{\ell_w(g,\dy)}.
\end{align}
Unfortunately, this is suboptimal since the error may not converge to zero even given infinite data:
\begin{proposition}\label{prop:loose}
	Consider the set the dynamics class $\calT=\{T_\theta:\theta\in S^{d-1},\theta_i\ge 0,\forall i\in[d]\}.$ Let $\Pi=\{\pi_x:x\in [d]\}$ where $\pi_x(s_i)=a_x$ for $0\le i\le d$ and $\pi_x(s_g)=\pi_x(s_b)=a_1.$ Let $\calW$ be the density ratio class induced by $\pi$ running on $\{T^\star\}\cup \mathcal{T}$. Even with $\calG=\{V^{\pi_x}_\dystar:x\in [d]\}$ and infinite number of data, we have
	\begin{align}
		\min_{\dy\in\caldy}\max_{w\in\calW,g\in\calG}\abs{\ell_w(g,\dy)}\ge \frac{\gamma}{8(1-\gamma)}.
	\end{align}
\end{proposition}
\noindent In contrast, the error terms in Theorem~\ref{thm:OPE-main} converge to $0$ when $\zeta>\poly(d,1/(1-\gamma))$ and $n\to\infty$ in the same setting.

\begin{proof}[Proof of Proposition~\ref{prop:loose}]
Recall that we set the dynamics class $\calT=\{T_\theta:\theta\in S^{d-1}\}.$ Let $\Pi=\{\pi_x:x\in [d]\}$ where $\pi_x(s_i)=a_x$ for $0\le i\le d$ and $\pi_x(s_g)=\pi_x(s_b)=a_1.$ Let $\calW$ be the density ratio induced by $\pi$. For any $x\in[d]$, we can compute 
\begin{align}
	&\rho^{\pi_x}_\dystar(s_0,a_i)=(1-\gamma)\ind{i=x},\quad \rho^{\pi_x}_\dystar(s_i,a_j)=\gamma(1-\gamma)\ind{i=x,j=x},\\
	&\rho^{\pi_x}_\dystar(s_g,a_j)=\gamma^2(1-\gamma)\ind{j=1},\quad \rho^{\pi_x}_\dystar(s_b,a_j)=0.
\end{align}
Let $\mu$ be uniform distribution over $3d+d^2$ state action pairs. Then we can define $\calW=\{w_x:x\in [d]\}$ where $w_x(s,a)\defeq \frac{1}{1-\gamma}\frac{\rho^{\pi_x}_\dystar(s,a)}{\mu(s,a)}.$ \

Now for any fixed $\theta\in S^{d-1},\theta\ge 0$, consider 
\begin{align}
	\max_{w\in\calW,g\in\calG}\abs{\ell_w(g,\dy_\theta)}.
\end{align}
Let $x=\argmin_{i}\theta_i.$ We claim that $$\ell_{w_x}(V^{\pi_x}_\dystar,\dy_\theta)\ge \frac{\gamma}{8(1-\gamma)}.$$
Indeed, with infinite data we have
\begin{align*}
	\ell_{w_x}(V^{\pi_x}_\dystar,\dy_\theta)&=\abs{\E_{(s,a)\sim \mu}\bbra{w_x(s,a)\pbra{\E_{s'\sim \dy(s,a)}[V^{\pi_x}_\dystar(s')]-\E_{s'\sim \dystar(s,a)}[V^{\pi_x}_{\dystar}(s')]}}}\\
	&=\frac{1}{1-\gamma}\abs{\E_{(s,a)\sim \rho^{\pi_x}_\dystar}\bbra{\pbra{\E_{s'\sim \dy(s,a)}[V^{\pi_x}_\dystar(s')]-\E_{s'\sim \dystar(s,a)}[V^{\pi_x}_{\dystar}(s')]}}}.
\end{align*}
Recall that $\dy_\theta=\dystar$ for states $s_0,s_g,s_b$. As a result, we continue the equation by
\begin{align*}
	&\frac{1}{1-\gamma}\abs{\E_{(s,a)\sim \rho^{\pi_x}_\dystar}\bbra{\pbra{\E_{s'\sim \dy(s,a)}[V^{\pi_x}_\dystar(s')]-\E_{s'\sim \dystar(s,a)}[V^{\pi_x}_{\dystar}(s')]}}}\\
	=\;&\gamma\abs{\E_{s'\sim \dy(s_x,a_x)}[V^{\pi_x}_\dystar(s')]-\E_{s'\sim \dystar(s_x,a_x)}[V^{\pi_x}_{\dystar}(s')]}\tag{by the definition of $\rho$}\\
	=\;&\gamma\abs{\frac{1}{2}(1+\theta_x)V^{\pi_x}_\dystar(s_g)+\frac{1}{2}(1-\theta_x)V^{\pi_x}_\dystar(s_b)-V^{\pi_x}_\dystar(s_g)}\tag{by the definition of $\dy_\theta$}\\
	=\;&\frac{\gamma}{2}\pbra{1-\theta_x}\pbra{V^{\pi_x}_\dystar(s_g)-V^{\pi_x}_\dystar(s_b)}.
\end{align*}
By basic algebra, $V^{\pi_x}_\dystar(s_g)=(1-\gamma)^{-1}$ and $V^{\pi_x}_\dystar(s_b)=0$. As a result, we get
\begin{align}
	\ell_{w_x}(V^{\pi_x}_\dystar,\dy_\theta)\ge \frac{\gamma}{2(1-\gamma)}(1-\theta_x).
\end{align}
Recall that $x=\argmin_{i}\theta_i$. Since $\theta\in S^{d-1}$ and $\theta_i\ge 0,\forall i$, we have $1=\sum_{i=1}^{d}\theta_i^2\ge d\theta_x^2.$ As a result, when $d>2$ we have $\theta_x\le 1/\sqrt{2}.$ Therefore
\begin{align}
    \ell_{w_x}(V^{\pi_x}_\dystar,\dy_\theta)\ge \frac{\gamma}{2(1-\gamma)}(1-\theta_x)\ge \frac{\gamma}{8(1-\gamma)}.
\end{align}
\end{proof}

\section{Helper Lemmas}
In this section, we present several helper lemmas used in Appendix~\ref{app:proofs}.
\begin{lemma}\label{lem:helper-TV-to-IS}
	For two distribution $p,q$ over $x\in\calX$, if we have $\normone{p-q}\le \epsilon$, then for any $\zeta>1$,
	$$\E_{x\sim p}\bbra{\ind{\frac{p(x)}{q(x)}>\zeta}}\le \frac{\zeta}{\zeta-1}\epsilon.$$
\end{lemma}
\begin{proof}
	Define $E(x)=\ind{\frac{p(x)}{q(x)}>\zeta}.$ Note that under event $E(x)$ we have
	\begin{align}
		p(x)>q(x)\zeta\implies p(x)-q(x)>q(x)(\zeta-1).
	\end{align}
	As a result, 
	\begin{align}
		\epsilon&\ge \normone{p-q}\ge \int\abs{p(x)-q(x)}E(x)\;dx\\
		&\ge \int(\zeta-1)q(x)E(x)\;dx=\E_{x\sim q}\bbra{E(x)}(\zeta-1)\\
		&\ge \pbra{\E_{x\sim p}\bbra{E(x)}-\epsilon}(\zeta-1).
	\end{align}
	By algebraic manipulation we get $\E_{x\sim p}\bbra{E(x)}\le \frac{\zeta}{\zeta-1}\epsilon.$
\end{proof}

\begin{lemma}\label{lem:TV-chainrule}
	Consider a fixed policy $\pi$ and two dynamics model $\dy,\bar{\dy}$. Suppose $$\E_{(s,a)\sim \rho^\pi_\dy}\bbra{\TV{\dy(s,a)}{\bar{\dy}(s,a)}}\le \epsilon,$$ we get
	\begin{equation}
		\normone{\rho^\pi_\dy-\rho^\pi_{\bar\dy}}\le \frac{1}{1-\gamma}\epsilon.
	\end{equation}
\end{lemma}
\begin{proof}
	First of all let $G,\bar{G}$ be the transition kernel from $\calS\times \calA$ to $\calS\times \calA$ induced by $\dy,\pi$ and $\bar{\dy},\pi$ respectively. Then for any distribution $\rho\in \Delta\pbra{\calS\times \calA}$ we have
	\begin{align}
		\normone{G\rho-\bar{G}\rho}\le \E_{(s,a)\sim \rho}\bbra{\TV{\bar\dy(s,a)}{\dy(s,a)}}.
	\end{align}
	Let $\rho_h$ (or $\bar{\rho}_h$) be the state-action distribution on step $h$ under dynamics $\dy$ (or $\bar\dy$). Then we have
	\begin{equation}
		\rho_h-\bar\rho_h=\pbra{G^h-\bar G^h}\rho_0=\sum_{h'=0}^{h-1}\bar G^{h-h'-1}\pbra{G-\bar G}G^{h'}\rho_0.
	\end{equation}
	As a result,
	\begin{align}
		&\normone{\rho_h-\bar\rho_h}\le \sum_{h'=0}^{h-1}\normone{\bar G^{h-h'-1}\pbra{G-\bar G}G^{h'}\rho_0}\\ 
		&\quad \le\sum_{h'=0}^{h-1}\normone{\pbra{G-\bar G}G^{h'}\rho_0}\le \sum_{h'=0}^{h-1}\E_{(s,a)\sim \rho_{h'}}\bbra{\TV{\bar\dy(s,a)}{\dy(s,a)}}.
	\end{align}
	It follows that 
	\begin{align}
		&\normone{\rho^\pi_\dy-\rho^\pi_{\bar\dy}}\le (1-\gamma)\sum_{h=0}^{\infty}\gamma^h\normone{\rho_h-\bar\rho_h}\\
		\le &(1-\gamma)\sum_{h=0}^{\infty}\gamma^h \sum_{h'=0}^{h-1}\E_{(s,a)\sim \rho_{h'}}\bbra{\TV{\bar\dy(s,a)}{\dy(s,a)}}\\
		\le &(1-\gamma)\sum_{h=0}^{\infty}\frac{\gamma^h}{1-\gamma}\E_{(s,a)\sim \rho_{h}}\bbra{\TV{\bar\dy(s,a)}{\dy(s,a)}}\\
		= &\sum_{h=0}^{\infty}\gamma^h\E_{(s,a)\sim \rho_{h}}\bbra{\TV{\bar\dy(s,a)}{\dy(s,a)}}\\
		= &\frac{1}{1-\gamma}\E_{(s,a)\sim \rho^\pi_\dy}\bbra{\TV{\bar\dy(s,a)}{\dy(s,a)}}.
	\end{align}
\end{proof}

\section{LQR Experimental Details}\label{sec:exp_appx}

\subsection{Data generation} 
The offline dataset is generated by running several $\pi_v$ under the true dynamics with $v\in\{-1, -0.75, -0.5, -0.25, 0, 0.25, 0.5, 0.75\}$ and added noise $\calN(0,0.5)$ to the policy. As a result, the behavior dataset covers most of the state-action space. The dataset contains $2000$ trajectories with length $20$ from each policy.

\subsection{Implementation} We compute the density ratio by approximating the behavior distribution $\mu$ and the state-action distribution $\rho^\pi_\dy$ respectively. By discretizing the state-action space into $10\times 10$ bins uniformly, the distribution $\mu(s,a)$ is approximated by the frequency of the corresponding bin. For $\rho^\pi_\dy$, we first collect $2000$ trajectories of policy $\pi$ under $\dy$ and compute the distribution similarly. Because all the function classes are finite, we enumerate over the function classes to compute $\lb(\dy,\pi)$ for every pair of dynamics and policy.

\subsection{Hyperparameters}
In the experiments, we use the following hyperparameters.
\begin{itemize}
	\item Cutoff threshold in Line 3 of Alg.~\ref{alg:OPE}: $\zeta=50.$
	\item Random seeds for three runs: 1, 2, 3.
	\item State noise: $\eta\sim \calN(0,0.05).$
	\item Policy noise: $\calN(0,0.01).$
	\item Discount factor: $\gamma=0.9.$
	\item Mean of initial state: $0.5$.
	\item Noise added to initial state: $0.2.$
	\item Number of trajectories per policy: $2000$.
\end{itemize}

We do not require parameter tuning for optimization procedures. We tried cutoff threshold with $\zeta\in\{10,20,50\}$ and number of trajectories in $\{20, 500, 2000\}.$ Smaller cutoff leads to an over-pessimistic lower bound, and fewer trajectories introduce variance to the final result.

\subsection{Computing resources}
These experiments run on a machine with 2 CPUs, 4GB RAM, and Ubuntu 20.04. We don't require GPU resources. We use Python 3.9.5 and numpy 1.20.2.

\section{D4RL Experimental Details}\label{sec:exp_appx_d4rl}

\subsection{Tasks}

\paragraph{Hopper.} The Hopper task is to make a hopper with three joints and four body parts hop forward as fast as possible. The state space is 11-dimension, the action is a 3-dimensional continuous space.

\paragraph{HalfCheetah.} The HalfCheetah task is to make a 2D robot with 7 rigid links, including 2 legs and a torso run forward as fast as possible. The state space is 17-dimension, the action is a 6-dimensional continuous space.

\subsection{Model Choice and Hyperparameters}
For all the dynamics, each model is parametrized as a 4-layer feedforward neural network with 200 hidden units. For the SAC~\citep{haarnoja2018} updates (serving as the policy gradient updates subroutine), the function approximations used for the policy and value function are 2-layer feedforward neural networks with 256 hidden units.

The hyperparameter choices for behavior density modeling are based on the training progress of the normalizing flow model. We pre-select a few (less than 10) combinations of hyperparameters and pick the set that gives us the lowest training loss. Usually, this is not the best practice. However, the small number of combinations (non-exhaustive search) and small model size reduced our concern for training set overfitting.

\paragraph{MOPO~\citep{yu2020mopo}:}
\begin{itemize}
    \item Batch size: 100.
    \item Rollout horizon: 5.
    \item Lambda: 1.
\end{itemize}

\paragraph{MBLB:}
\begin{itemize}
    \item Random seeds for five runs: 1, 2, 3, 4, 5.
    \item Number of trajectories to sample: 100.
    \item Rollout horizon: 5.
    \item Batch size: 32.
    \item Cutoff threshold in Line 3 of Alg.~\ref{alg:OPE}: $\zeta=5.$
    \item Discount factor $\gamma$: 0.99.
    \item GAE $\lambda$: 0.95.
    \item $g$ function latent size: 8.
\end{itemize}

\paragraph{MML:}

\begin{itemize}
    \item Random seeds for five runs: 1, 2, 3, 4, 5.
    \item Batch size: 32.
    \item Basis function class: square, polynomial
    \item Ratio-Value function parametrization: linear, reproducing kernel hilbert space (RKHS) 
\end{itemize}

For MML, we first need to make a decision on how to parametrize $h(s, a, s')$. If we choose a linear parametrization such as $h(s, a, s') = \psi(s, a, s')^T \theta$, we need to decide what $\psi$ is. There are two obvious choices: $\psi(x) = [x, x^2, 1]$ (\textbf{square} basis function), or a \textbf{polynomial} basis function with degree 2: given $x = [x_1, x_2, ..., x_d]$, $\psi(x) = [x_1^2, x_1 x_2, x_1 x_3, ..., x_2^2, x_2 x_3, ..., x_d^2]$, which can be efficiently computed as the upper triangular entries of $x x^T$.  If we choose the ratio-value function parametrization to be RKHS, then we use radial basis function (RBF) as $K((s, a, s'), (\tilde s, \tilde a, \tilde s'))$.

\subsection{Computing resources}
These experiments run on a machine with 4 CPUs, 10GB RAM, and Ubuntu 20.04. We don't require GPU resources. We use Python 3.9.5 and numpy 1.20.2.

\section{Algorithms}

We describe the MML and MBLB algorithms in this section. Algorithm~\ref{alg:mblb} describes how we compute MBLB. Note that we compute three components of lower bound explicitly. Algorithm~\ref{alg:mml_linear} describes how we compute MML with linear parametrization. Algorithm~\ref{alg:mml_kernel} describes how we compute MML with RKHS parametrization.

\noindent\makebox[\textwidth][c]{
\begin{minipage}{0.55\textwidth}
\centering
\begin{algorithm}[H]
\SetAlgoLined
\SetKwFunction{fqe}{trainFQE}
\SetKwFunction{flow}{trainFlow}
\SetKwFunction{init}{Initialize}
\SetKwFunction{sample}{Sample}
\SetKwFunction{append}{append}
\KwIn{offline RL data $\mathcal{D}$; set of dynamics, policy pairs [$(\pi_1, T_1), ..., (\pi_K, T_K)$], $V_{\max}$, $\gamma$, $\zeta$.}
\KwOut{optimal policy $\pi^*$}
\BlankLine
$\hat\mu(\cdot, \cdot)$ = \flow($\mathcal{D}$)\\
scores = [] \\
\For{$i \leftarrow 1...K$}{
    $Q^{\pi_i}$ = \fqe(\sample($\mathcal{D}$, $T_i$, $\pi_i$), $\pi_i$) \\
    $\rho_{\pi_i}^{T_i}(\cdot, \cdot)$ = \flow(\sample($\mathcal{D}, T_i, \pi_i$))\\
    $\eta = \mathbb{E}_{(s, a) \sim \mathcal{D}}[Q^{\pi_i}(s, \pi_i(s))]$ \\
    \init($\theta$) \\
    $L = 0; \Delta = 0$ \\
    \For{$(s, a, s') \in \mathcal{D}$}{
        $w = \max(\min(\frac{\rho^{T_i}_{\pi_i}(s, a)}{\hmu(s, a)}, \zeta), 0)$ \\ %
        $\ell = - |w \cdot (\mathbb{E}_{x \sim T_i(s)}[g_\theta(x)] - g_\theta(s'))|$ \\
        $\theta = \theta + \nabla_\theta \ell$ \\
        $\Delta = \Delta - V_{\max} \cdot \ind{\frac{\rho^{T_i}_{\pi_i}(s, a)}{\hmu(s, a)} > \zeta}$ \\
        $L = L + \ell$ \\
    }
    score = $\frac{1}{|\mathcal{D}|}(\eta + \frac{1}{1-\gamma} (\Delta + L))$ \\
    scores $\leftarrow$ score
} 
 $i = \argmax($scores$)$ \\
 \Return{$\pi_i$}
 \caption{MBLB: Model-based Lower Bound}
 \label{alg:mblb}
\end{algorithm}
\end{minipage}
}

\noindent\makebox[\textwidth][c]{
\begin{minipage}{0.55\textwidth}
\centering
\begin{algorithm}[H]
\SetAlgoLined
\SetKwFunction{init}{Initialize}
\SetKwFunction{append}{append}
\KwIn{offline RL data $\mathcal{D}$; set of dynamics, policy pairs [$(\pi_1, T_1), ..., (\pi_K, T_K)$].}
\KwOut{optimal policy $\pi^*$}
\BlankLine
scores = [] \\
\For{$i \leftarrow 1...K$}{
    \init($\theta$) \\
    $L = 0$ \\
    \For{$(s, a, s') \in \mathcal{D}$}{
        $\ell = - (\mathbb{E}_{x \sim T_i(s)}[\psi(s, a, x)^T \theta] - \psi(s, a, s')^T \theta)$ \\
        $\theta = \theta + \nabla_\theta \ell$ \\
        $L = L + \ell$ \\
    }
    score = $\frac{L}{|\mathcal{D}|}$ \\
    scores $\leftarrow$ score
} 
 $i = \argmax($scores$)$ \\
 \Return{$\pi_i$}
 \caption{MML-Linear: Minimax Model Learning Bound}
 \label{alg:mml_linear}
\end{algorithm}
\end{minipage}
}

\noindent\makebox[\textwidth][c]{
\begin{minipage}{0.55\textwidth}
\centering
\begin{algorithm}[H]
\SetAlgoLined
\SetKwFunction{subsample}{Subsample}
\SetKwFunction{fqe}{trainFQE}
\SetKwFunction{init}{Initialize}
\SetKwFunction{append}{append}
\KwIn{offline RL data $\mathcal{D}$; set of dynamics, policy pairs [$(\pi_1, T_1), ..., (\pi_K, T_K)$], kernel $K$.}
\KwOut{optimal policy $\pi^*$}
\BlankLine
scores = [] \\
\For{$i \leftarrow 1...K$}{
    $L = 0$ \\
    \For{$(s, a, s'), (\tilde s, \tilde a, \tilde s') \in \mathcal{D}$}{
        $\ell_1 = \mathbb{E}_{x \sim T(s), \tilde x \sim T(\tilde s)}[K((s, a, x), (\tilde s, \tilde a, \tilde x))]$ \\
        $\ell_2 = - 2 \mathbb{E}_{x \sim T(s)}[K((s, a, x), (\tilde s, \tilde a, \tilde s'))]$ \\
        $\ell_3 = K((s, a, s'), (\tilde s, \tilde a, \tilde s'))$ \\
        $L = L + \ell_1 + \ell_2 + \ell_3$ \\
    }
    score = $\frac{L}{|\mathcal{D}|}$ \\
    scores $\leftarrow$ score
} 
 $i = \argmax($scores$)$ \\
 \Return{$\pi_i$}
 \caption{MML-RKHS: Minimax Model Learning Bound}
 \label{alg:mml_kernel}
\end{algorithm}
\end{minipage}
}

\section{D4RL Additional Experiments}\label{sec:add_exp_appx_d4rl}

\subsection{Ablation Study}
We conduct an ablation study in Table~\ref{tab:fqe_d4rl} where we evaluate the final performance of the policies selected using either FQE with TD-1 estimation or FQE with GAE estimation. We observe that using GAE for offline policy selection allows for picking better policies on average.

\begin{table*}[]
\centering
\begin{tabular}{@{}lccc@{}}
\toprule
Dataset Type & Environment &  \begin{tabular}[c]{@{}c@{}}FQE\\ (TD-1)\end{tabular}       & \begin{tabular}[c]{@{}c@{}}FQE\\ (GAE)\end{tabular}                \\ \midrule
medium       & hopper      & \begin{tabular}[c]{@{}c@{}}507.8 \\ (549.6)\end{tabular}   & \begin{tabular}[c]{@{}c@{}}533.5 \\ (532.6)\end{tabular}                  \\ \midrule
med-expert   & hopper      & \begin{tabular}[c]{@{}c@{}}149.3 \\ (146.2)\end{tabular}   & \begin{tabular}[c]{@{}c@{}}261.1 \\ (157.9)\end{tabular}         \\ \midrule
expert       & hopper      & \begin{tabular}[c]{@{}c@{}}39.0 \\ (34.6)\end{tabular}     & \begin{tabular}[c]{@{}c@{}}120.7 \\ (78.7)\end{tabular}        \\ \midrule
medium       & halfcheetah & \begin{tabular}[c]{@{}c@{}}1802.5 \\ (1011.9)\end{tabular} & \begin{tabular}[c]{@{}c@{}}2117.4 \\ (1215.6)\end{tabular}                    \\ \midrule
med-expert   & halfcheetah & \begin{tabular}[c]{@{}c@{}}302.1 \\ (605.2)\end{tabular}   & \begin{tabular}[c]{@{}c@{}}394.9 \\ (632.0)\end{tabular}       \\

\bottomrule
\end{tabular}
\caption{We report the mean and (standard deviation) of the selected policy's environment performance across 3 random seeds using different variants of FQE.}
\label{tab:fqe_d4rl}
\end{table*}

\subsection{MBLB with RKHS}
In this section, we derive the closed-form solution to $\sup_{g\in\calG}\ell_w(g,T)$ when the test function $g$ belongs to a reproducing kernel Hilbert space (RKHS), and empirically evaluate the MBLB method with RKHS parameterization.

Let $K:\calS\times\calS\to \R$ be a symmetric and positive definite kernel and $\calH_K$ its corresponding RKHS with inner product $\dotp{\cdot}{\cdot}_{\calH_K}$. Then we have the following lemma.

\begin{lemma} When $\calG=\{g\in \calH_K: \dotp{g}{g}_{\calH_K}\le 1\}$, we have
\begin{align}
    \sup_{g\in\calG}\ell_w(g,T)^2=\E_{s,a,s'\sim \calD,x\sim T(s,a)}\E_{\tilde{s},\tilde{a},\tilde{s}'\sim \calD,\tilde{x}\sim T(\tilde{s},\tilde{a})}\left[w(s,a)w(\tilde{s},\tilde{a})(K(x,\tilde{x})+K(s',\tilde{s}')-K(x,\tilde{s}')-K(\tilde{x},s')\right]
\end{align}
\end{lemma}
\begin{proof}
    Let $K_x\defeq K(x,\cdot)\in\calH_K$. By the reproducing property, we have $\dotp{K_x}{K_y}_{\calH_K}=K(x,y)$ and $\dotp{K_x}{g}_{\calH_K}=g(x)$. As a result,
    \begin{align}
        &\sup_{g\in\calG}\ell_w(g,T)^2=\sup_{g:\dotp{g}{g}_{\calH_K}\le 1}\E_{s,a,s'\sim \calD,x\sim T(s,a)}[w(s,a)(\dotp{K_x}{g}_{\calH_K}-\dotp{K_{s'}}{g}_{\calH_K})]^2\\
        =&\;\sup_{g:\dotp{g}{g}_{\calH_K}\le 1}\dotp{\E_{s,a,s'\sim \calD,x\sim T(s,a)}[w(s,a)(K_x-K_{s'})]}{g}_{\calH_K}^2\\
        =&\;\|\E_{s,a,s'\sim \calD,x\sim T(s,a)}[w(s,a)(K_x-K_{s'})]\|_{\calH_K}^2\tag{Cauchy-Schwarz}\\
        =&\;\dotp{\E_{s,a,s'\sim \calD,x\sim T(s,a)}[w(s,a)(K_x-K_{s'})]}{\E_{\tilde{s},\tilde{a},\tilde{s}'\sim \calD,\tilde{x}\sim T(\tilde{s},\tilde{a})}[w(\tilde s,\tilde a)(K_{\tilde x}-K_{\tilde s'})]}_{\calH_K}\\
        =&\;\E_{s,a,s'\sim \calD,x\sim T(s,a)}\E_{\tilde{s},\tilde{a},\tilde{s}'\sim \calD,\tilde{x}\sim T(\tilde{s},\tilde{a})}[\dotp{w(s,a)(K_x-K_{s'})}{w(\tilde s,\tilde a)(K_{\tilde x}-K_{\tilde s'})}_{\calH_K}]\\
        =&\;\E_{s,a,s'\sim \calD,x\sim T(s,a)}\E_{\tilde{s},\tilde{a},\tilde{s}'\sim \calD,\tilde{x}\sim T(\tilde{s},\tilde{a})}[w(s,a)w(\tilde s,\tilde a)(K(x,\tilde{x})+K(s',\tilde{s}')-K(x,\tilde{s}')-K(\tilde{x},s')].
    \end{align}
\end{proof}

Table~\ref{tab:mblb-rkhs} presents the performance of the MBLB algorithm with RKHS parameterization. On most of the environments, MBLB-RKHS performs better than/comparable with MML-RKHS. However, MBLB-Quad consistently outperforms MBLB-RKHS on all the environments. We suspect that MBLB-RKHS could outperform MBLB-Quad with different choices of kernels because the quadratic parameterization can be seen as a special case of RKHS parameterization (with quadratic kernels).

\begin{table*}[htp]
\centering
\small
\begin{tabular}{@{}lcccccccc@{}}
\toprule
Dataset Type & Env & MOPO                                                     & \begin{tabular}[c]{@{}c@{}}MML\\ (Squared)\end{tabular}             & \begin{tabular}[c]{@{}c@{}}MML\\ (Polynomial)\end{tabular} & \multicolumn{1}{l}{\begin{tabular}[c]{@{}c@{}}MML\\ (RKHS)\end{tabular}} & \begin{tabular}[c]{@{}c@{}} \textbf{MBLB}\\ \textbf{(Linear)}\end{tabular} & \begin{tabular}[c]{@{}c@{}} \textbf{MBLB}\\ \textbf{(Quad)} \end{tabular} &  \begin{tabular}[c]{@{}c@{}} \textbf{MBLB}\\ \textbf{(RKHS)}\end{tabular} \\ \midrule
medium       & hopper      & \begin{tabular}[c]{@{}c@{}}175.4 \\ (95.3)\end{tabular}  & \begin{tabular}[c]{@{}c@{}}379.4 \\ (466.4)\end{tabular}            & \begin{tabular}[c]{@{}c@{}}375.6 \\ (459.5)\end{tabular}   & \begin{tabular}[c]{@{}c@{}}375.0 \\ (459.9)\end{tabular}                & \begin{tabular}[c]{@{}c@{}}591.7 \\ (523.1)\end{tabular} &    \begin{tabular}[c]{@{}c@{}}\textbf{808.5} \\ (502.7)\end{tabular} & \begin{tabular}[c]{@{}c@{}}317.8 \\ (476.4)\end{tabular}     \\ \midrule
med-expert   & hopper      & \begin{tabular}[c]{@{}c@{}}183.8 \\ (94.4)\end{tabular}  & \begin{tabular}[c]{@{}c@{}}160.9 \\ (131.5)\end{tabular}            & \begin{tabular}[c]{@{}c@{}}116.5 \\ (148.4)\end{tabular}  & \begin{tabular}[c]{@{}c@{}}61.4 \\ (35.0)\end{tabular}                  & \begin{tabular}[c]{@{}c@{}}\textbf{261.1} \\ (157.9)\end{tabular} &   \begin{tabular}[c]{@{}c@{}}242.5 \\ (134.0)\end{tabular} & \begin{tabular}[c]{@{}c@{}}208.1 \\ (144.3)\end{tabular}     \\ \midrule
expert       & hopper      & \begin{tabular}[c]{@{}c@{}}80.4 \\ (63.4)\end{tabular}   & \begin{tabular}[c]{@{}c@{}}93.8 \\ (87.9)\end{tabular}              & \begin{tabular}[c]{@{}c@{}}61.6 \\ (61.9)\end{tabular}    & \begin{tabular}[c]{@{}c@{}}70.0 \\ (56.2)\end{tabular}                  & \begin{tabular}[c]{@{}c@{}} 118.2 \\ (61.6)\end{tabular} &     \begin{tabular}[c]{@{}c@{}}\textbf{121.0} \\ (72.5)\end{tabular} &  \begin{tabular}[c]{@{}c@{}}120.9 \\ (61.8)\end{tabular}    \\ \midrule
medium       & halfcheetah & \begin{tabular}[c]{@{}c@{}}599.8 \\ (668.4)\end{tabular} & \begin{tabular}[c]{@{}c@{}}1967.6 \\ (1707.5)\end{tabular} & \begin{tabular}[c]{@{}c@{}}2625.1 \\ (937.2)\end{tabular} & \textbf{\begin{tabular}[c]{@{}c@{}}3858.2 \\ (1231.1)\end{tabular}}     &  \begin{tabular}[c]{@{}c@{}}3290.4 \\ (1753.1)\end{tabular} &       \begin{tabular}[c]{@{}c@{}}2484.2 \\ (1526.8)\end{tabular} &   \begin{tabular}[c]{@{}c@{}}2229.7 \\ (1949.8)\end{tabular}      \\ \midrule
med-expert   & halfcheetah & \begin{tabular}[c]{@{}c@{}}-486.6 \\ (48.1)\end{tabular} & \begin{tabular}[c]{@{}c@{}}-188.5 \\ (137.2)\end{tabular}            & \begin{tabular}[c]{@{}c@{}}-77.0 \\ (252.5)\end{tabular}   & \begin{tabular}[c]{@{}c@{}}-343.2\\ (225.2)\end{tabular}                 & \begin{tabular}[c]{@{}c@{}} \textbf{207.4} \\ (509.5) \end{tabular} &  \begin{tabular}[c]{@{}c@{}} 192.8 \\ (432.0) \end{tabular}  & \begin{tabular}[c]{@{}c@{}} -2.1 \\ (690.6) \end{tabular}   \\
\bottomrule
\end{tabular}
\caption{We report the mean and (standard deviation) of selected policy's simulator environment performance across 5 random seeds. MML and MBLB are used as model-selection procedures where they select the best policy for each seed. Our method is choosing the most near-optimal policy across the datasets.}
\label{tab:mblb-rkhs}
\end{table*}

\begin{figure*}[htp]
\centering
	\includegraphics[width=0.5\linewidth]{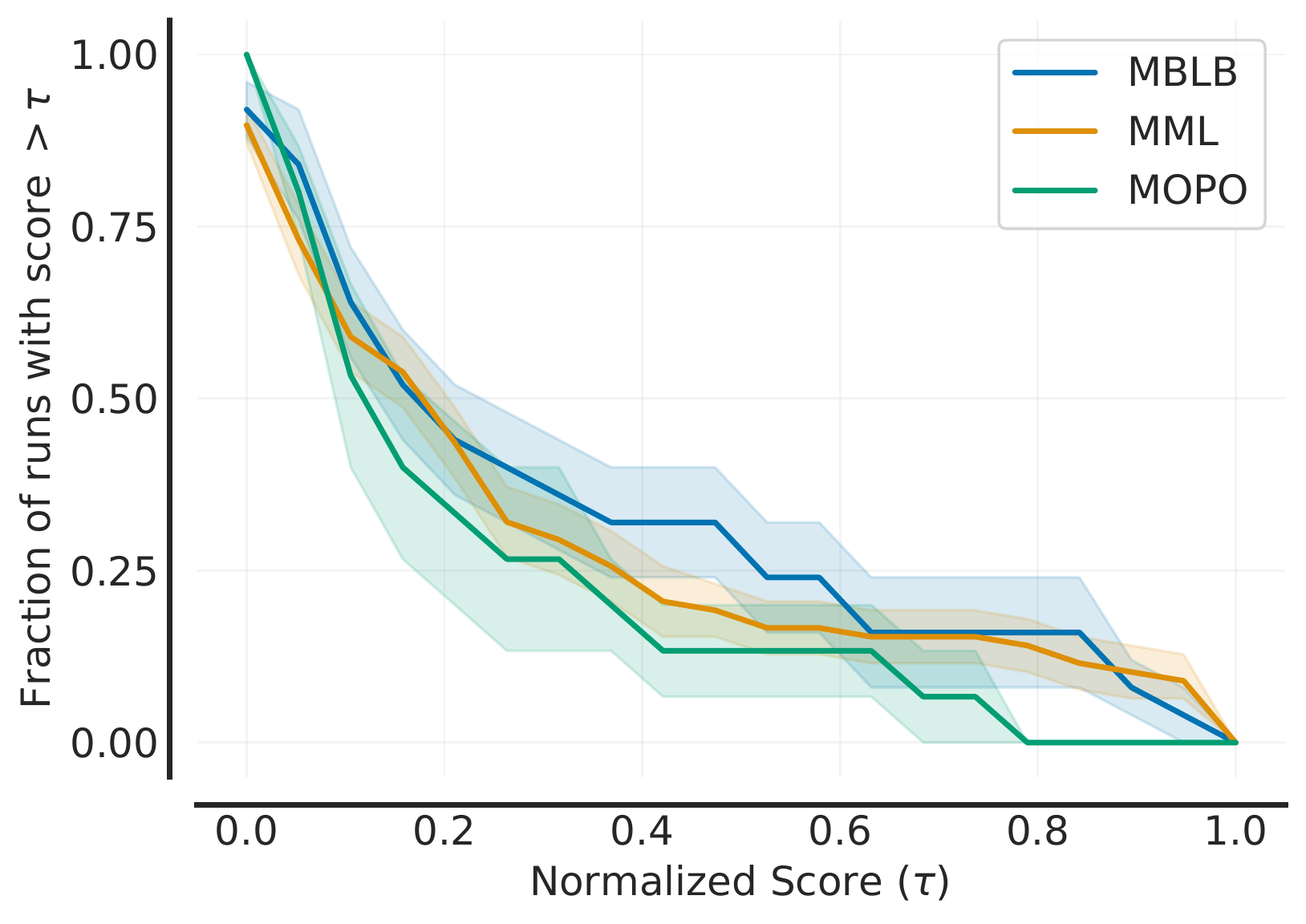}
\caption{Performance profile between three methods.  %
}
\label{fig:perf_profile}
\end{figure*}

\end{document}